\documentclass[10pt,journal,compsoc]{IEEEtran}
\ifCLASSOPTIONcompsoc
  \usepackage[nocompress]{cite}
\else
  \usepackage{cite}
\fi
\ifCLASSINFOpdf
\else
\fi
\usepackage{epsfig}
\usepackage{graphicx}
\graphicspath{{Figures/}}
\usepackage{amsmath,amsfonts, amssymb}
\usepackage{relsize}
\usepackage{tabularx}
\newcolumntype{Y}{>{\centering\arraybackslash}X}

\usepackage{multirow}
\usepackage{blindtext}
\usepackage{booktabs}
\usepackage{mmstyles}
\usepackage{bm}
\usepackage{diagbox}
\usepackage[skip=2pt]{caption}
\usepackage[percent]{overpic}
\usepackage{enumitem}
\usepackage{url}
\usepackage[table, dvipsnames]{xcolor}
\usepackage{colortbl}
\usepackage{tabularx}
\definecolor{Gray}{gray}{0.95}

\definecolor{todo}{rgb}{0.6,0,0} 

\definecolor{update}{rgb}{0,0,0.8} 
\definecolor{revise}{rgb}{0,0,0} \newcommand{\revise}[1]{\textcolor{revise}{#1}}

\makeatletter
\DeclareRobustCommand\onedot{\futurelet\@let@token\@onedot}
\def\@onedot{\ifx\@let@token.\else.\null\fi\xspace}

\def\eg{\emph{e.g}\onedot} 
\def\ie{\emph{i.e}\onedot}

\def\etal{\emph{et al}\onedot}
\makeatother
\begin{document}
%
% paper title
% Titles are generally capitalized except for words such as a, an, and, as,
% at, but, by, for, in, nor, of, on, or, the, to and up, which are usually
% not capitalized unless they are the first or last word of the title.
% Linebreaks \\ can be used within to get better formatting as desired.
% Do not put math or special symbols in the title.
\title{Old Photo Restoration via Deep Latent Space Translation}
\author{Ziyu Wan,
    Bo Zhang,
    Dongdong Chen,
    Pan Zhang,
    Dong Chen,
    Jing Liao,
    Fang Wen% <-this % stops a space
    \IEEEcompsocitemizethanks{\IEEEcompsocthanksitem Z. Wan is with the Department
        of Computer Science, City University of Hong Kong, Hong Kong SAR, China.\protect\\
        E-mail: ziyuwan2-c@my.cityu.edu.hk
        \IEEEcompsocthanksitem B. Zhang, D. Chen and F. Wen are with the Visual
        Computing Group, Microsoft Research, Beijing, China.\protect\\
        E-mail: \{Tony.Zhang, doch, fangwen\}@microsoft.com
        \IEEEcompsocthanksitem D. Chen is with Microsoft Cloud+AI, Redmond, Washington, USA.\protect\\
        E-mail: cddlyf@gmail.com
        \IEEEcompsocthanksitem P. Zhang is with the Department of Automation, University of Science and Technology of China, Hefei, Anhui, China.\protect\\
        E-mail: zhangpan@mail.ustc.edu.cn
        \IEEEcompsocthanksitem J. Liao is with the Department of Computer Science, City University of Hong Kong, Hong Kong SAR, China.\protect\\
        E-mail: jingliao@cityu.edu.hk
        \IEEEcompsocthanksitem J. Liao is the corresponding author.
    }% <-this % stops a space
    \thanks{Manuscript received April 19, 2005; revised August 26, 2015.}}

% note the % following the last \IEEEmembership and also \thanks - 
% these prevent an unwanted space from occurring between the last author name
% and the end of the author line. i.e., if you had this:
% 
% \author{....lastname \thanks{...} \thanks{...} }
%                     ^------------^------------^----Do not want these spaces!
%
% a space would be appended to the last name and could cause every name on that
% line to be shifted left slightly. This is one of those "LaTeX things". For
% instance, "\textbf{A} \textbf{B}" will typeset as "A B" not "AB". To get
% "AB" then you have to do: "\textbf{A}\textbf{B}"
% \thanks is no different in this regard, so shield the last } of each \thanks
% that ends a line with a % and do not let a space in before the next \thanks.
% Spaces after \IEEEmembership other than the last one are OK (and needed) as
% you are supposed to have spaces between the names. For what it is worth,
% this is a minor point as most people would not even notice if the said evil
% space somehow managed to creep in.

% The paper headers
\markboth{Journal of \LaTeX\ Class Files,~Vol.~14, No.~8, August~2015}%
{Shell \MakeLowercase{\textit{et al.}}: Bare Advanced Demo of IEEEtran.cls for IEEE Computer Society Journals}
% The only time the second header will appear is for the odd numbered pages
% after the title page when using the twoside option.
% 
% *** Note that you probably will NOT want to include the author's ***
% *** name in the headers of peer review papers.                   ***
% You can use \ifCLASSOPTIONpeerreview for conditional compilation here if
% you desire.

% The publisher's ID mark at the bottom of the page is less important with
% Computer Society journal papers as those publications place the marks
% outside of the main text columns and, therefore, unlike regular IEEE
% journals, the available text space is not reduced by their presence.
% If you want to put a publisher's ID mark on the page you can do it like
% this:
%\IEEEpubid{0000--0000/00\$00.00~\copyright~2015 IEEE}
% or like this to get the Computer Society new two part style.
%\IEEEpubid{\makebox[\columnwidth]{\hfill 0000--0000/00/\$00.00~\copyright~2015 IEEE}%
%\hspace{\columnsep}\makebox[\columnwidth]{Published by the IEEE Computer Society\hfill}}
% Remember, if you use this you must call \IEEEpubidadjcol in the second
% column for its text to clear the IEEEpubid mark (Computer Society journal
% papers don't need this extra clearance.)

% use for special paper notices
%\IEEEspecialpapernotice{(Invited Paper)}

% for Computer Society papers, we must declare the abstract and index terms
% PRIOR to the title within the \IEEEtitleabstractindextext IEEEtran
% command as these need to go into the title area created by \maketitle.
% As a general rule, do not put math, special symbols or citations
% in the abstract or keywords.
\IEEEtitleabstractindextext{%
    \begin{abstract}
        We propose to restore old photos that suffer from severe degradation through a deep learning approach. Unlike conventional restoration tasks that can be solved through supervised learning, the degradation in real photos is complex and the domain gap between synthetic images and real old photos makes the network fail to generalize. Therefore, we propose a novel triplet domain translation network by leveraging real photos along with massive synthetic image pairs. Specifically, we train two variational autoencoders (VAEs) to respectively transform old photos and clean photos into two latent spaces. And the translation between these two latent spaces is learned with synthetic paired data. This translation generalizes well to real photos because the domain gap is closed in the compact latent space. Besides, to address multiple degradations mixed in one old photo, we design a global branch with a partial nonlocal block targeting to the structured defects, such as scratches and dust spots, and a local branch targeting to the unstructured defects, such as noises and blurriness. Two branches are fused in the latent space, leading to improved capability to restore old photos from multiple defects. Furthermore, we apply another face refinement network to recover fine details of faces in the old photos, thus ultimately generating photos with enhanced perceptual quality. With comprehensive experiments, the proposed pipeline demonstrates superior performance over state-of-the-art methods as well as existing commercial tools in terms of visual quality for old photos restoration.
    \end{abstract}

    % Note that keywords are not normally used for peerreview papers.
    }

% make the title area
\maketitle

% To allow for easy dual compilation without having to reenter the
% abstract/keywords data, the \IEEEtitleabstractindextext text will
% not be used in maketitle, but will appear (i.e., to be "transported")
% here as \IEEEdisplaynontitleabstractindextext when compsoc mode
% is not selected <OR> if conference mode is selected - because compsoc
% conference papers position the abstract like regular (non-compsoc)
% papers do!
\IEEEdisplaynontitleabstractindextext
% \IEEEdisplaynontitleabstractindextext has no effect when using
% compsoc under a non-conference mode.

% For peer review papers, you can put extra information on the cover
% page as needed:
% \ifCLASSOPTIONpeerreview
% \begin{center} \bfseries EDICS Category: 3-BBND \end{center}
% \fi
%
% For peerreview papers, this IEEEtran command inserts a page break and
% creates the second title. It will be ignored for other modes.
\IEEEpeerreviewmaketitle

\ifCLASSOPTIONcompsoc
    \IEEEraisesectionheading{\section{Introduction}\label{sec:introduction}}
\else
    \section{Introduction}
    \label{sec:introduction}
\fi

\IEEEPARstart{P}{hotos} are taken to freeze the happy moments that otherwise gone. Even though time goes by, one can still evoke memories of the past by viewing them. Nonetheless, old photo prints deteriorate when kept in poor environmental condition, which causes the valuable photo content permanently damaged. Fortunately, as mobile cameras and scanners become more accessible, people can now digitalize the photos and invite a skilled specialist for restoration. However, manual retouching is usually laborious and time-consuming, which leaves piles of old photos impossible to get restored. Hence, it is appealing to design automatic algorithms that can instantly repair old photos for those who wish to bring old photos back to life.

Prior to the deep learning era, there are some attempts~\cite{stanco2003towards,bruni2004generalized,chang2005photo,giakoumis2005digital} that restore photos by automatically detecting the localized defects such as scratches and blemishes, and filling in the damaged areas with inpainting techniques. Yet these methods focus on completing the missing content and none of them can repair the spatially-uniform defects such as film grain, sepia effect, color fading, etc., so the photos after restoration still appear outdated compared to modern photographic images. With the emergence of deep learning, one can address a variety of low-level image restoration problems~\cite{zhang2017learning,zhang2017beyond,dong2014learning,xu2014deep,ren2016single,zhang2019deep,gao2019deep} by exploiting the powerful representation capability of convolutional neural networks, \ie, learning the mapping for a specific task from a large amount of synthetic images.

The same framework, however, does not apply to old photo restoration and the reason is three-fold. First, the degradation process of old photos is rather complex, and there exists no degradation model that can realistically render the old photo artifact. Therefore, the model learned from those synthetic data generalizes poorly on real photos. Second, old photos are plagued with a compound of degradation and inherently require different strategies for repair: unstructured defects that are spatially homogeneous, \eg, film grain and color fading, should be restored by utilizing the pixels in the neighborhood, whereas the structured defects, \eg, scratches, dust spots, etc., should be repaired with a global image context. \revise{Furthermore, people are fastidious to tiny artifacts around faces yet a network trained on general natural images cannot capture facial intrinsic characteristics. Thus, a network targeting for face retouching is needed especially considering portraits account for large proportion of old photos.}

\begin{figure*}[t!]
    \begin{center}
        \includegraphics[width=1.0\linewidth]{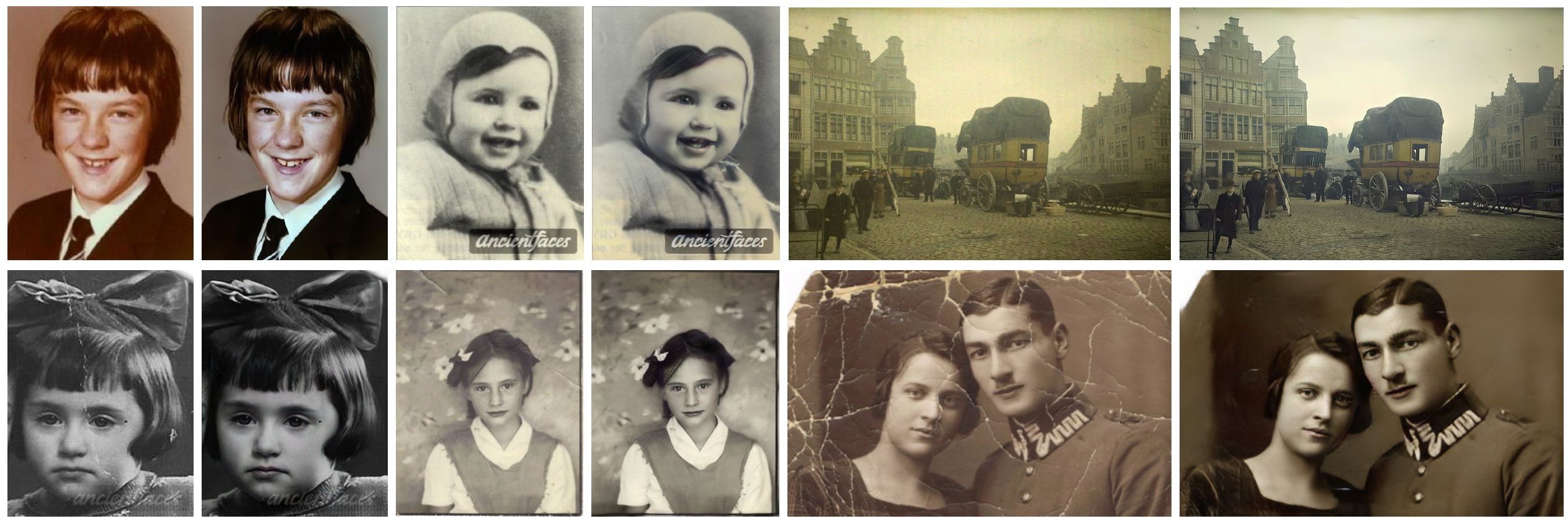}
    \end{center}
    % \vspace{-1em}
    \caption{\textbf{Old photo restoration results produced by our method.} Our method can handle the complex degradation mixed
        by both unstructured and structured defects in real old photos. In particular, we recover high-frequency details for face regions, further improving the perceptual quality for portraits. For each image pair, left is the input while the retouched output is shown on the right.}
    \label{figure:qualitative_show}
    %\vspace{-0.6em}
\end{figure*}

To circumvent these issues, we formulate the old photo restoration as a triplet domain translation problem. Different from previous image translation methods~\cite{isola2017image}, we leverage data from three domains (\ie, real old photos, synthetic images and the corresponding ground truth), and the translation is performed in latent space. Synthetic images and the real photos are first transformed to the same latent space with a shared variational autoencoder~\cite{kingma2013auto} (VAE). Meanwhile, another VAE is trained to project ground truth clean images into the corresponding latent space. The mapping between the two latent spaces is then learned with the synthetic image pairs, which restores the corrupted images to clean ones. The advantage of the latent restoration is that the learned latent restoration can generalize well to real photos because of the domain alignment within the first VAE. Besides, we differentiate the mixed degradation and propose a partial nonlocal block that considers the long-range dependencies of latent features to specifically address the structured defects during the latent translation. \revise{Finally, considering that faces are the most important visual stimuli, we propose a post-processing step with a coarse-to-fine generator to reconstruct high-resolution faces with hierarchical spatial adaptive conditions. Some results are shown in Figure~\ref{figure:qualitative_show}.} In comparison with several leading restoration methods, we prove the effectiveness of our approach in restoring multiple degradations of real photos.

\section{Related Work}
\label{sec:relateG_work}
\noindent\textbf{Single degradation image restoration.} ~~
%According to the complexity of corruption, we categorize image restoration tasks into two groups. The first group is to remove an unstructured defect such as noise, blurriness, color fading, or low resolution, corresponding to the methods of denoising, deblurring, recoloring, and super resolution. The second group requires higher-level understanding of the image to repair the structures of the image in larger-size damaged areas such as holes, scratches, or spots, known as inpainting. In the past decades, extensive research works have defined different priors to tackle each single restoration problem in either group. Recently, deep learning has demonstrated its superior ability in single degradation image restoration tasks by utilizing large-scale training data. We briefly review the related works below. 
Existing image degradation can be roughly categorized into two groups: unstructured degradation such as noise, blurriness, color fading, and low resolution, and structured degradation such as holes, scratches, and spots. For the former unstructured ones, traditional works often impose different image priors, including nonlocal self-similarity~\cite{buades2005non,mairal2009non,dabov2007image}, sparsity~\cite{elad2006image,mairal2007sparse,yang2010image,xie2012image} and local smoothness~\cite{weiss2007makes,babacan2008total,li2009markov}. Recently, a lot of deep learning based methods have also been proposed for different image degradation, like image denoising~\cite{zhang2017learning,zhang2017beyond,zhang2018ffdnet,mao2016image,lefkimmiatis2018universal,liu2018non,zhang2019rnan}, super-resolution~\cite{dong2014learning,kim2016accurate,ledig2017photo,wang2018esrgan,zhang2018residual}, and deblurring~\cite{xu2014deep,sun2015learning,nah2017deep,kupyn2018deblurgan}.

Compared to unstructured degradation, structured degradation is more challenging and often modeled as the ``image painting" problem. Thanks to powerful semantic modeling ability, most existing best-performed inpainting methods are learning based. For example, Liu \etal~\cite{liu2018image} masked out the hole regions within the convolution operator and enforces the network focus on non-hole features only.  To get better inpainting results, many other methods consider both local patch statistics and global structures. Specifically, Yu \etal~\cite{yu2018generative} and Liu \etal~\cite{liu2019coherent} proposed to employ an attention layer to utilize the remote context. And the appearance flow is explicitly estimated by Ren \etal~\cite{ren2019structureflow} so that textures in the hole regions can be directly synthesized based on the corresponding patches.

No matter for unstructured or structured degradation, though the above learning-based methods can achieve remarkable results, they are all trained on the synthetic data. Therefore, their performance on the real dataset highly relies on synthetic data quality. For real old images, since they are often seriously degraded by a mixture of unknown degradation, the underlying degradation process is much more difficult to be accurately characterized. In other words, the network trained on synthetic data only, will suffer from the domain gap problem and perform badly on real old photos. In this paper, we model real old photo restoration as a new triplet domain translation problem and some new techniques are adopted to minimize the domain gap.

\noindent\textbf{Mixed degradation image restoration.} ~~
In the real world, a corrupted image may suffer from complicated defects mixed with scratches, loss of resolution, color fading, and film noises. However, research solving mixed degradation is much less explored. The pioneer work RL-Restore~\cite{yu2018crafting} proposed a toolbox that comprises multiple light-weight networks, and each of them responsible for a specific degradation. Then they learn a controller that dynamically selects the operator from the toolbox. Inspired by RL-Restore~\cite{yu2018crafting}, Suganum \etal~\cite{suganuma2018attention} performs different convolutional operations in parallel and uses the attention mechanism to select the most suitable combination of operations. However, these methods still rely on supervised learning from synthetic data and hence cannot generalize to real photos. Besides, they only focus on unstructured defects and do not support structured defects like image inpainting. On the other hand, DIP~\cite{ulyanov2018deep} found that the deep neural network inherently resonates with low-level image statistics and thereby can be utilized as an image prior for blind image restoration without external training data. This method has the potential, though not claimed in DIP~\cite{ulyanov2018deep}, to restore in-the-wild images corrupted by mixed factors. In comparison, our approach excels in both restoration performance and efficiency.

\noindent\textbf{Face restoration.}
\revise{
    A variety of methods specifically designed for face restoration have been proposed. Early works~\cite{hacohen2013deblurring,pan2014deblurring} attempt to deblur faces by the guidance of an external reference, but an exemplar image with suitable texture for transfer is inconvenient to retrieve and the requirement of an external face database makes it cumbersome for practical usage. On the other hand, most contemporary works~\cite{chan2019everybody} rely on generative adversarial network (GAN) to resolve the blurriness and produce realistic result. It is noteworthy that the restoration quality could be boosted by explicitly considering intrinsic facial priors such as face parsing~\cite{shen2018deep}, facial landmarks~\cite{bulat2018super}, identity prior~\cite{grm2019face} or 3D morphable models~\cite{ren2019face}. Nonetheless, these methods require extra networks to perform those auxiliary tasks, which brings robustness issue when processing the face images that suffer from large pose and severe degradations. A recent work~\cite{menon2020pulse} utilizes a pre-trained generative model and searches the latent code that conforms to the input. Albeit impressive, the generated faces suffer from fidelity issue. In this work, we aim to restore in-the-wild faces with well-preserved identity while caring for robustness. To this end, we do not rely on face prior and learn the restoration by synthesis: instead of letting the network digest the degraded faces as input, the output is synthesized from a latent noise with the latent features modulated by the degraded faces through spatially-variant de-normalization. We will show that this approach achieves preferable quality in restoring vintage portraits.}

\noindent\textbf{Old photo restoration.}~~
Old photo restoration is a classical mixed degradation problem, but most existing methods ~\cite{stanco2003towards,bruni2004generalized,chang2005photo,giakoumis2005digital} focus on inpainting only. They follow a similar paradigm \ie, defects like scratches and blotches are first identified according to low-level features and then inpainted by borrowing the textures from the vicinity. However, the hand-crafted models and low-level features they used are difficult to detect and fix such defects well. Moreover, none of these methods consider restoring some unstructured defects such as color fading or low resolution together with inpainting. Thus photos still appear old fashioned after restoration. In this work, we reinvestigate this problem by virtue of a data-driven approach, which can restore images from multiple defects simultaneously and turn heavily-damaged old photos to modern style.

\section{Method}
% \subsection{Overview}
In contrast to conventional image restoration tasks, old photo restoration is more challenging. First, old photos contain far more complex degradation that is hard to be modeled realistically and there always exists a domain gap between synthetic and real photos. As such, the network usually cannot generalize well to real photos by purely learning from synthetic data. Second, the defects of old photos is a compound of multiple degradations, thus essentially requiring different strategies for restoration. Unstructured defects such as film noise, blurriness and color fading, etc. can be restored with spatially homogeneous filters by making use of surrounding pixels within the local patch; structured defects such as scratches and blotches, on the other hand, should be inpainted by considering the global context to ensure the structural consistency. \revise{In the following, we first describe our main framework to address the aforementioned \emph{generalization issue} and \emph{mixed degradation issue} respectively. After that, we introduce auxiliary network for face enhancement, so as to further improve the restoration quality.}

\subsection{Restoration via latent space translation}
In order to mitigate the domain gap, we formulate the old photo restoration as an image translation problem, where we treat clean images and old photos as images from distinct domains and we wish to learn the mapping in between. However, as opposed to general image translation methods that bridge two different domains~\cite{isola2017image, CycleGAN}, we translate images across three domains: the real photo domain $\cR$, the synthetic domain $\cX$ where images suffer from artificial degradation, and the corresponding ground truth domain $\cY$ that comprises images without degradation. Such triplet domain translation is crucial in our task as it leverages the unlabeled real photos as well as a large amount of synthetic data associated with ground truth.

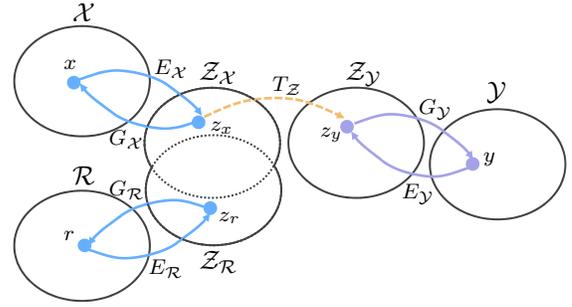
\begin{figure}[t!]
    \centering
    \small
    \begin{overpic}
        % [scale=0.4,grid,tics=5]{Figures/diagram4.pdf} 
        [scale=0.4]{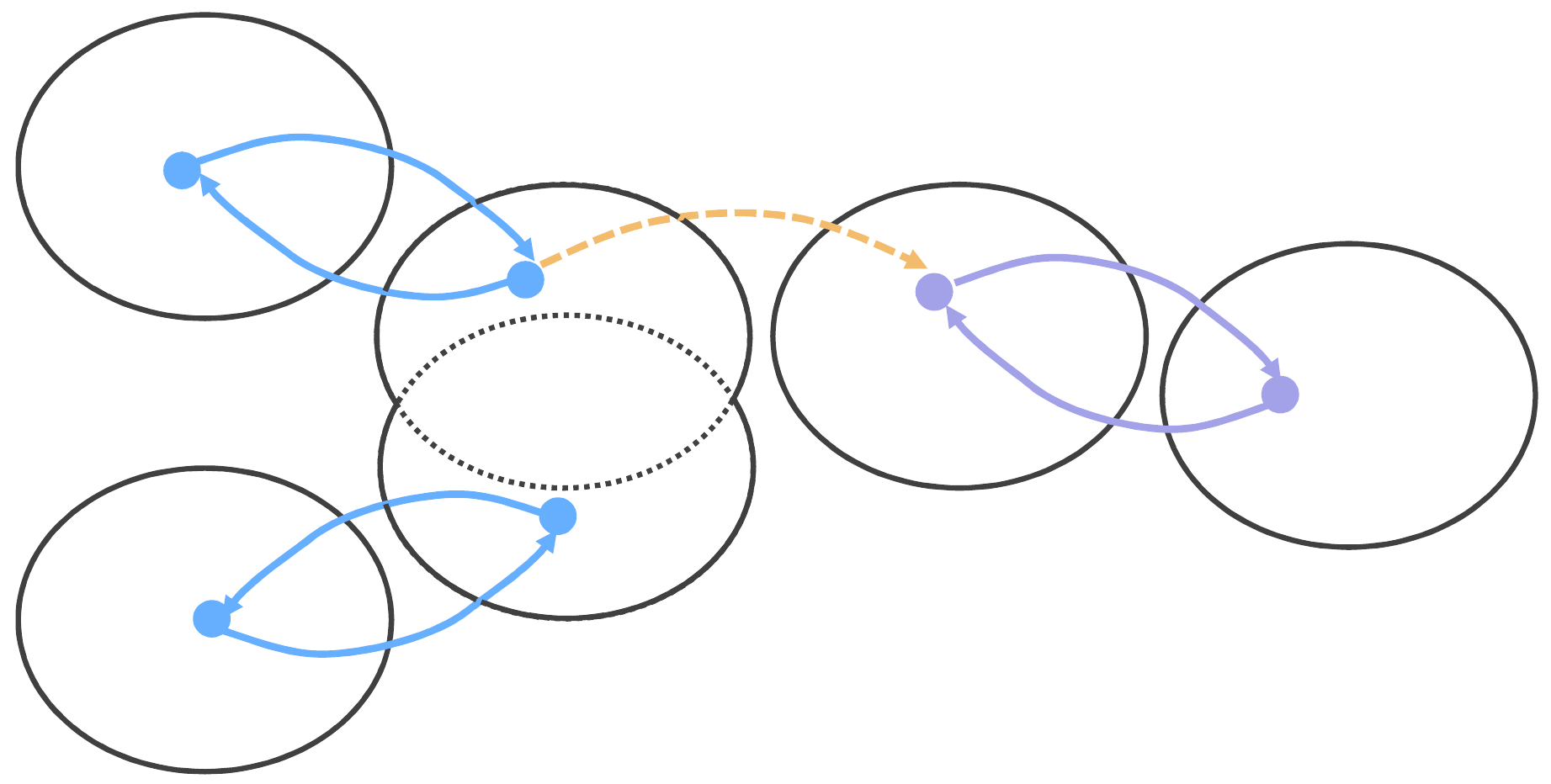}
        \put(10,42){\footnotesize$x$}
        \put(10,11.5){\footnotesize$r$}
        \put(84.5,25.5){\footnotesize$y$}
        \put(36,31){\footnotesize$z_x$}
        \put(37.5,15){\footnotesize$z_r$}
        \put(55.5,30){\footnotesize$z_y$}
        \put(11.5,50.5){\normalsize$\cX$}
        \put(11.5,21.5){\normalsize$\cR$}
        \put(34,40){\normalsize$\cZ_\cX$}
        \put(34,6.5){\normalsize$\cZ_{\cR}$}
        \put(60,40){\normalsize$\cZ_\cY$}
        \put(85,36.5){\normalsize$\cY$}
        \put(26,41){\footnotesize$E_{\cX}$}
        \put(18,19.5){\footnotesize$G_{\cR}$}
        \put(18,28.5){\footnotesize$G_{\cX}$}
        \put(25,5.5){\footnotesize$E_{\cR}$}
        \put(47,38){\footnotesize$T_{\cZ}$}
        \put(73,34){\footnotesize$G_{\cY}$}
        \put(70,19){\footnotesize$E_{\cY}$}
    \end{overpic}
    % \caption{\textbf{Illustration of our translation method with three domains.} We use one VAE to project the images from domain $\cX$ and $\cR$ into a shared latent space. Then another VAE is trained to project clean images $\cY$ into $\cZ_{\cY}$. Based on the synthetic pair relations, the latent space translation $T_{\cZ}$ will be learned between $\cZ_{\cX}$ and $\cZ_{\cY}$.}
    \caption{\textbf{Illustration of our translation method with three domains.} The domain gap between $\cZ_{\cX}$ and $\cZ_{\cR}$ will be reduced in the shared latent space.}
    % %\vspace{-1.5em}
    \label{fig:diagram1}
\end{figure}

We denote images from three domains respectively with  $r\in \cR$, $x\in \cX$ and $y\in \cY$, where $x$ and $y$ are paired by data synthesis, \ie, $x$ is degraded from $y$. Directly learning the mapping from real photos $\{r\}_{i=1}^N$ to clean images $\{y\}_{i=1}^N$ is hard since they are not paired and thus unsuitable for supervised learning. We thereby propose to decompose the translation with two stages, which are illustrated in Figure~\ref{fig:diagram1}. First, we propose to map $\cR$, $\cX$, $\cY$ to corresponding latent spaces via $E_{\cR}:\cR \mapsto \cZ_{\cR}$, $E_{\cX}:\cX \mapsto \cZ_{\cX}$, and $E_{\cY}:\cY \mapsto \cZ_{\cY}$, respectively. In particular, because synthetic images and real old photos are both corrupted, sharing similar appearances, we align their latent space into the shared domain by enforcing some constraints. Therefore we have $\cZ_{\cR} \approx \cZ_{\cX}$. This aligned latent space encodes features for all the corrupted images, either synthetic or real ones. Then we propose to learn the image restoration in the latent space. Specifically, by utilizing the synthetic data pairs~$\{x,y\}_{i=1}^N$, we learn the translation from the latent space of corrupted images, $\cZ_{\cX}$, to the latent space of ground truth, $\cZ_{\cY}$, through the mapping $T_{\cZ}:\cZ_{\cX} \mapsto \cZ_{\cY}$, where $\cZ_{\cY}$ can be further reversed to $\cY$ through generator $G_{\cY}:{\cZ_{\cY}} \mapsto \cY$. By learning the latent space translation, real old photos $r$ can be restored by sequentially performing the mappings,
% \vspace{-0.65em}
\begin{equation}
    r_{\cR \to \cY} =  G_{\cY} \circ T_{\cZ} \circ E_{\cR}(r).
\end{equation}

% \begin{figure}[t!]
%     % \centering
%     \hspace{1em}
%     \small
%     \begin{overpic}
%         % [scale=0.6,grid,tics=5]{Figures/diagram3.pdf} 
%         [scale=0.55]{Figures/diagram3.pdf} 
%         \put(-2,73.5){$r$}
%         \put(-2,69.5){$x$}
%         \put(-2,15){$y$}
%         \put(99,73.5){${r}_{\cR \to \cR}$}
%         \put(99,69.5){${x}_{\cX \to \cR}$}
%         \put(99,21){${r}_{\cR \to \cY}$}
%         \put(99,17){${x}_{\cX \to \cY}$}
%         \put(99,13){${y}_{\cY \to \cY}$}
%         \put(15,71){\textcolor{white}{$E_{\cR,\cX}$}}
%         \put(17,13){\textcolor{white}{$E_\cY$}}
%         \put(75,71){\textcolor{white}{$G_{\cR,\cX}$}}
%         \put(77,13){\textcolor{white}{$G_\cY$}}
%         \put(48,69){$z_\cR, z_\cX$}
%         \put(48,16.5){$z_\cY$}
%     \end{overpic}
%     \caption{\textbf{Illustration of the restoration network.} We first train a VAE-GAN model for the corrupted and clean domain, respectively. In each domain, images can be  represented by their corresponding latent code. Then,
%     we learn the mapping that restore the corrupted images to clean ones in the latent space. (draw procedures)}
%     \label{fig:diagram}
% \end{figure}

\begin{figure}[t!]
    % \centering
    \hspace{1em}
    \small
    \begin{overpic}
        % [scale=0.55,grid,tics=5]{Figures/diagram5.pdf} 
        [scale=0.58]{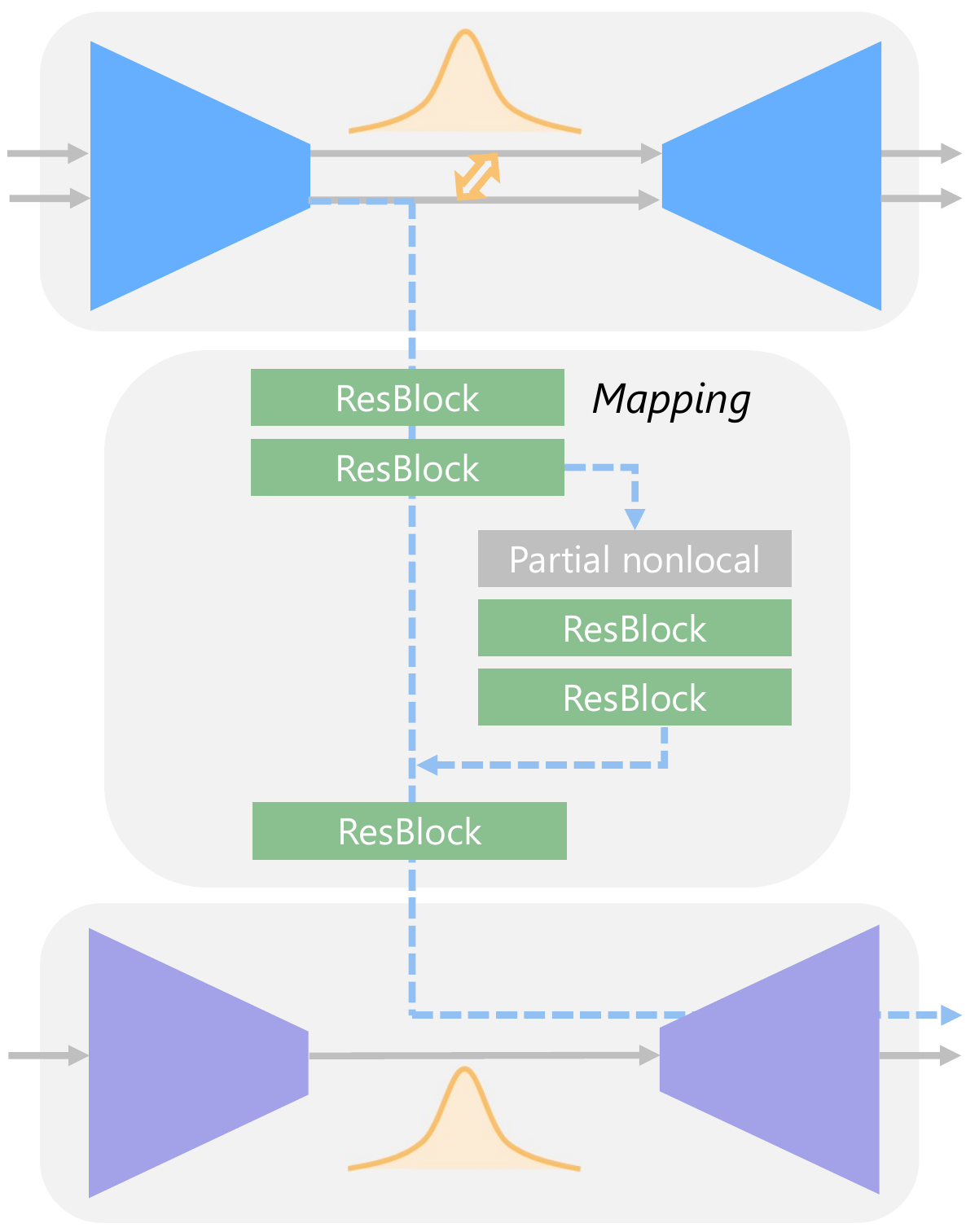}
        \put(-2,86.5){$r$}
        \put(-2,83.5){$x$}
        \put(-2,14.5){$y$}
        \put(78,86.5){${r}_{\cR \to \cR}$}
        \put(78,83.5){${x}_{\cX \to \cX}$}
        \put(78,21){${r}_{\cR \to \cY}$}
        \put(78,17){${x}_{\cX \to \cY}$}
        \put(78,13){${y}_{\cY \to \cY}$}
        \put(11.5,84){\textcolor{white}{$E_{\cR,\cX}$}}
        \put(12.5,13){\textcolor{white}{$E_\cY$}}
        \put(59,84){\textcolor{white}{$G_{\cR,\cX}$}}
        \put(60.5,13){\textcolor{white}{$G_\cY$}}
        \put(35,80){$z_\cR, z_\cX$}
        \put(35,15.5){$z_\cY$}
        \put(35,19){$z_{\cR\to\cY},z_{\cX\to\cY}$}
        \put(42,93){$\cN(0,I)$}
        \put(42,9){$\cN(0,I)$}
        \put(42,84.5){$adv.$}
        \put(15,95.5){I.}
        \put(15,23){I.}
        \put(15,66){II.}
        \put(61.5,66.2){$\cT$}
    \end{overpic}
    \caption{\textbf{Architecture of our restoration network.} (I.) We first train two VAEs: VAE$_1$ for images in real photos $r\in\cR$ and synthetic images $x\in\cX$, with their domain gap closed by jointly training an adversarial discriminator; VAE$_2$ is trained for clean images $y\in\cY$. With VAEs, images are transformed to  compact latent space. (II.) Then,
        we learn the mapping that restores the corrupted images to clean ones in the latent space. }
    % \vspace{-3em}
    \label{fig:diagram_new}
\end{figure}
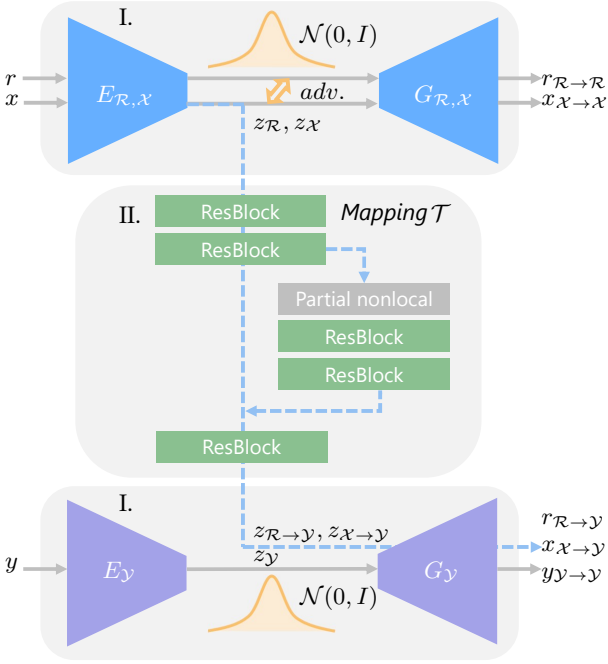

\noindent\textbf{Domain alignment in the VAE latent space}~~ One key of our method is to meet the assumption that ${\cR}$ and ${\cX}$ are encoded into the same latent space. To this end, we propose to utilize variational autoencoder~\cite{kingma2013auto} (VAE) to encode images with compact representation, whose domain gap is further examined by an adversarial discriminator~\cite{GAN}. We use the network architecture shown in Figure~\ref{fig:diagram_new} to realize this concept.

In the first stage, two VAEs are learned for the latent representation. Old photos $\{r\}$ and synthetic images $\{x\}$ share the first one termed VAE$_1$, with the encoder $E_{\cR,\cX}$ and generator $G_{\cR,\cX}$, while the ground true images $\{y\}$ are fed into the second one, VAE$_2$ with the encoder-generator pair $\{E_{\cY},G_{\cY}\}$. VAE$_1$ is shared for both $r$ and $x$ in the aim that images from both corrupted domains can be mapped to a shared latent space. The VAEs assumes Gaussian prior for the distribution of latent codes, so that images can be reconstructed by sampling from the latent space. We use the re-parameterization trick to enable differentiable stochastic sampling~\cite{KingmaW13} and optimize VAE$_1$ with data $\{r\}$ and $\{x\}$ respectively. The objective with $\{r\}$ is defined as:
\begin{equation}
    \begin{split}
        \cL_{\text{VAE}_1}(r) &=\ {\mathrm{KL}}(E_{\cR,\cX}(z_r|r) || \cN({0},{I})) \\ &+ \alpha\Ebb_{z_r\sim E_{\cR,\cX}(z_r|r)}\left[\norm{ G_{\cR,\cX}(r_{\cR\to\cR}|z_r)-r}_1\right] \\&+ \cL_{\text{VAE}_1,\text{GAN}}(r)
    \end{split}
    \label{eq:vae1}
\end{equation}
%and, 
%\begin{equation}
%    \begin{split}
%    \cL_{\text{VAE}_1}(x) =\ &{\mathrm{KL}}(E_{\cR,\cX}(z_x|x) || \cN({0},{I})) \\ &+ \alpha\Ebb_{z_x\sim E_{\cR,\cX}(z_x|x)}\left[\norm{ G_{\cR,\cX}(x_{\cX\to\cX}|z_x)-x}_1\right] \\&+ \cL_{\text{VAE}_1,\text{GAN}}(x),
%\end{split}
%\label{eq:vae2}
%\end{equation}
where, $z_r\in\cZ_\cR$ is the latent codes for $r$, and $r_{\cR\to\cR}$ is the generation outputs. The first term in equations is the KL-divergence that penalizes deviation of the latent distribution from the Gaussian prior. The second $\ell_1$ term lets the VAE reconstruct the inputs, implicitly enforcing latent codes to capture the major information of images. Besides, we introduce the least-square loss (LSGAN)~\cite{mao2017least}, denoted as $\cL_{\text{VAE}_1,\text{GAN}}$ in the formula, to address the well-known over-smooth issue in VAEs, further encouraging VAE to reconstruct images with high realism. The objective with $\{x\}$, denoted as $\cL_{\text{VAE}_1}(x)$, is defined similarly. And VAE$_2$ for domain $\cY$ is trained with a similar loss so that the corresponding latent representation~$z_{y}\in \cY$ can be derived.

We use VAE rather than vanilla autoencoder because VAE features denser latent representation due to the KL regularization (which will be proved in ablation study), and this helps produce closer latent space for $\{r\}$ and $\{x\}$ with VAE$_1$ thus leading to smaller domain gap. To further narrow the domain gap in this reduced space, we propose to use an adversarial network to examine the residual latent gap. Concretely, we train another discriminator $D_{\cR,\cX}$ that differentiates $\cZ_\cR$ and $\cZ_\cX$, whose loss is defined as,
\begin{equation}
    \begin{split}
        \cL_{\text{VAE}_1,\text{GAN}}^{\text{latent}}(r,x) =\ &\Ebb_{x\sim \cX}[D_{\cR,\cX}(E_{\cR,\cX}(x))^2] \\ &+ \Ebb_{r\sim \cR}[(1-D_{\cR,\cX}(E_{\cR,\cX}(r)))^2].
    \end{split}
\end{equation}
Meanwhile, the encoder $E_{\cR,\cX}$ of VAE$_1$ tries to fool the discriminator with a contradictory loss to ensure that $\cR$ and $\cX$ are mapped to the same space. Combined with the latent adversarial loss, the total objective function for VAE$_1$ becomes,
\begin{equation}
    \min_{E_{\cR,\cX},G_{\cR,\cX}}\max_{D_{\cR,\cX}}\ \cL_{\text{VAE}_1}(r) + \cL_{\text{VAE}_1}(x) + \cL_{\text{VAE}_1,\text{GAN}}^{\text{latent}}(r,x).
\end{equation}

% The drawback of VAE is that the reconstructed results may lose some high frequency details. To alleviate this point, we also involve adversarial loss into the training process. Here we use LSGANs~\cite{mao2017least} directly, that is,
% \begin{equation}
% \mathcal{L}_{GAN} = \mathbb{E}[G_1(x)^2] + \mathbb{E}[(1-G_1(G_1(z)))^2].
% \end{equation}

\noindent\textbf{Restoration through latent mapping}~~
With the latent code captured by VAEs, in the second stage, we leverage the synthetic image pairs $\{x,y\}$ and propose to learn the image restoration by mapping their latent space (the mapping network $M$ in Figure~\ref{fig:diagram_new}). The benefit of latent restoration is threefold. First, as $\cR$ and $\cX$ are aligned into the same latent space, the mapping from $\cZ_{\cX}$ to $\cZ_{\cY}$ will also generalize well to restoring the images in $\cR$. Second, the mapping in a compact low-dimensional latent space is in principle much easier to learn than in the high-dimensional image space. In addition, since the two VAEs are trained independently and the reconstruction of the two streams would not be interfered with each other. The generator $G_{\cY}$ can always get an absolutely clean image without degradation given the latent code $z_{\cY}$ mapped from~$\cZ_{\cX}$, whereas degradations will likely remain if we learn the translation in pixel level.

Let $r_{\cR\to \cY}$, $x_{\cX\to \cY}$ and $y_{\cY \to \cY}$ be the final translation outputs for $r$, $x$ and $y$, respectively. At this stage, we solely train the parameters of the latent mapping network $\cT$ and fix the two VAEs. The loss function $\cL_{\cT}$, which is imposed at both the latent space and the end of generator $G_{\cY}$, consists of three terms,
\begin{equation}
    \cL_{\cT}(x,y) = \lambda_{1}\cL_{\cT,\ell_1} + \cL_{\cT,\text{GAN}} + \lambda_{2}\cL_{\text{FM}}
    \label{eq:translation}
\end{equation}
where the latent space loss, $\cL_{\cT,\ell_1} = \Ebb\norm{\cT(z_x)-z_y)}_1$, penalizes the $\ell_1$ distance of the corresponding latent codes. We introduce the adversarial loss $\cL_{\cT,\text{GAN}}$, still in the form of LSGAN~\cite{mao2017least}, to encourage the ultimate translated synthetic image $x_{\cX\to \cY}$ to look real. Besides, we introduce feature matching loss $L_{\text{FM}}$ to stabilize the GAN training. Specifically, $L_{\text{FM}}$ matches the multi-level activations of the adversarial network $D_{M}$, and that of the pretrained VGG network (also known as perceptual loss in~\cite{isola2017image,johnson2016perceptual}), \ie,
\begin{align}
    \mathcal{L}_{\text{FM}}
    % _{\text{FM}}&(x_{\cX\to \cY},y_{\cY\to \cY}) \\
    =\  & \Ebb\ \Big[\sum_i \frac 1 {n_{D_{\cT}}^i}\| \phi_{D_{\cT}}^i (x_{\cX\to \cY}) - \phi_{D_{\cT}}^i (y_{\cY\to \cY})\|_1 \nonumber \\
    +   & \sum_i \frac 1 {n_{\text{VGG}}^i} \| \phi_{\text{VGG}}^i (x_{\cX\to \cY}) - \phi_{\text{VGG}}^i (y_{\cY\to \cY})\|_1\Big],
\end{align}
where $\phi^i_{D_{\cT}}$ ($\phi^i_{\text{VGG}}$) denotes the $i^{th}$ layer feature map of the discriminator (VGG network), and $n^i_{D_{\cT}}$ ($n^i_{\text{VGG}}$)
indicates the number of activations in that layer.

% Regarding the implementation of $\cT$, three convolutional layers are first employed to expand the $z_x$ into the 512-channel feature maps. Symmetrically, in the end part three convolutions are used to map the feature maps into latent space $z_y$ (explanation?). Six residual blocks are inserted in between to enhance the translation capacity.

\subsection{Multiple degradation restoration }
\begin{figure}[tb!]
    \begin{center}
        \includegraphics[width=1.0\linewidth]{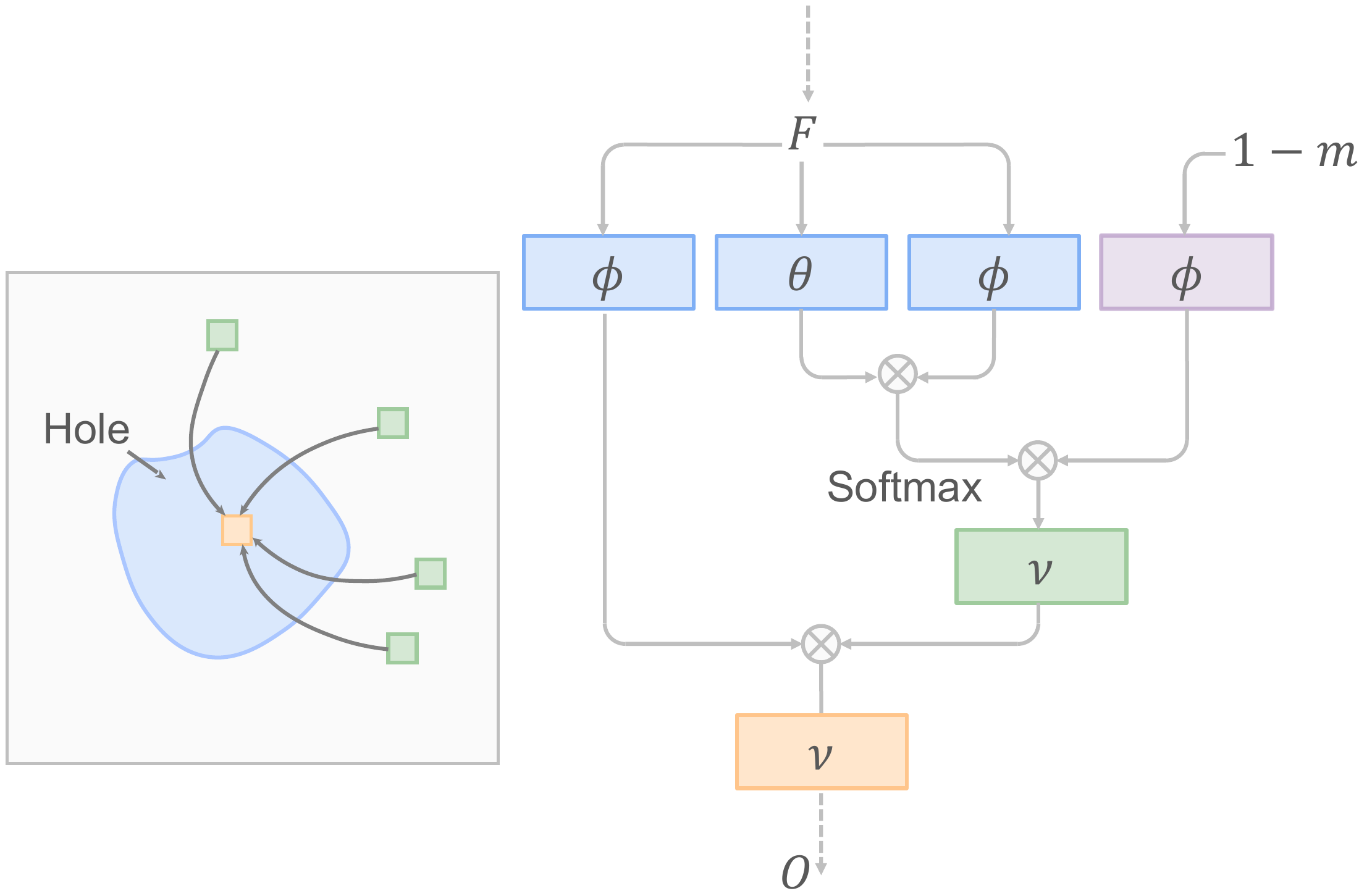}
    \end{center}
    \caption{\textbf{Partial nonlocal block}. Left shows the principle. The pixels within the hole areas are inpainted by the context pixels outside the corrupted region. Right shows the detailed implementation.}
    \label{figure:partial_nonlocal}
\end{figure}

The latent restoration using the residual blocks, as described earlier, only concentrates on local features due to the limited receptive field of each layer. Nonetheless, the restoration of structured defects requires plausible inpainting, which has to consider long-range dependencies so as to ensure global structural consistency. Since legacy photos often contain mixed degradations, we have to design a restoration network that simultaneously supports the two mechanisms. Towards this goal, we propose to enhance the latent restoration network by incorporating a global branch as shown in Figure~\ref{fig:diagram_new}, which composes of a nonlocal block~\cite{wang2018non} that considers global context and several residual blocks in the following. While the original block proposed in~\cite{wang2018non} is unaware of the corruption area, our nonlocal block explicitly utilizes the mask input so that the pixels in the corrupted region will not be adopted for completing those area. Since the context considered is a part of the feature map, we refer to the module specifically designed for the latent inpainting as a \emph{partial nonlocal block}, which is shown in Figure~\ref{figure:partial_nonlocal}.

Formally, let $F\in \Rbb^{C\times HW}$ be the intermediate feature map in $M$ ($C$, $H$ and $W$ are number of channels, height and width respectively), and $m\in \{0,1\}^{HW}$ represents the binary mask downscaled to the same size, where $1$ represents the defect regions to be inpainted and $0$ represents the intact regions. The affinity between $i^{th}$ location and $j^{th}$ location in $F$, denoted by $s_{i,j}\in\Rbb^{HW\times HW}$, is calculated by the correlation of $F_i$ and $F_j$ modulated by the mask $(1-m_j)$, \ie,
\begin{equation}
    s_{i,j}=(1-m_j)f_{i,j} / \sum_{\forall k} (1-m_k)f_{i,k},
\end{equation}
where,
\begin{equation}
    f_{i,j} = \exp(\theta(F_i)^T\cdot \phi(F_j))
\end{equation}
gives the pairwise affinity with embedded Gaussian. Here, $\theta$ and $\phi$ project $F$ to Gaussian space for affinity calculation. According to the affinity $s_{i,j}$ that considers the holes in the mask, the partial nonlocal finally outputs
\begin{equation}
    O_i = {\nu}\left(\sum_{\forall j} s_{i,j} {\mu}(F_j)\right),
\end{equation}
which is a weighted average of correlated features for each position. We implement the embedding functions ${\theta}$, ${\phi}$, ${\mu}$ and ${\nu}$ with 1$\times$1 convolutions.

We design the global branch specifically for inpainting and hope the non-hole regions are left untouched, so we fuse the global branch with the local branch under the guidance of the mask, \ie,
\begin{equation}
    F_{fuse} = (1-m)\odot \rho_{\text{local}}(F) + m\odot \rho_{\text{global}}(O),
\end{equation}
where operator $\odot$ denotes Hadamard product, and $\rho_{\text{local}}$ and $\rho_{\text{global}}$ denote the nonlinear transformation of residual blocks in two branches. In this way, the two branches constitute the latent restoration network, which is capable to deal with multiple degradation in old photos. We will detail the derivation of the defect mask in Section~\ref{sec:implementation}. 

Table~\ref{table:network} shows the detailed network structure.
\label{Network Architectures}
\begin{table}[!tb]
    \footnotesize
    \centering
    \begin{tabularx}{\linewidth}{Y|Y|Y|Y}
        \toprule
        Module                           & Layer                              & Kernel size / stride         & Output size                           \\ \midrule
        \multirow{4}{*}{Encoder $E$}     & Conv                               & $7\times7/1$                 & $256\times256\times64$                \\
                                         & Conv                               & $4\times4/2$                 & $128\times128\times64$                \\
                                         & Conv                               & $4\times4/2$                 & $64\times64\times64$                  \\
                                         & ResBlock$\times$4                  & $3\times3/1$                 & $64\times64\times64$                  \\ \midrule
        \multirow{4}{*}{Generator $G$}   & ResBlock$\times$4                  & $3\times3/1$                 & $64\times64\times64$                  \\
                                         & Deconv                             & $4\times4/2$                 & $128\times128\times64$                \\
                                         & Deconv                             & $4\times4/2$                 & $256\times256\times64$                \\
                                         & Conv                               & $7\times7/1$                 & $256\times256\times3$                 \\ \midrule
        \multirow{8}{*}{{Mapping $\cT$}} & Conv                               & $3\times3/1$                 & $64\times64\times128$                 \\
                                         & Conv                               & $3\times3/1$                 & $64\times64\times256$                 \\
                                         & Conv                               & $3\times3/1$                 & $64\times64\times512$                 \\
        \cmidrule{2-4}

                                         & \cellcolor{Gray} Partial nonlocal & \cellcolor{Gray}$1\times1/1$ & \cellcolor{Gray}$64\times64\times512$ \\

                                         & \cellcolor{Gray}Resblock$\times$2  & \cellcolor{Gray}$3\times3/1$ & \cellcolor{Gray}$64\times64\times512$ \\
        \cmidrule{2-4}
                                         & ResBlock$\times$6                  & $3\times3/1$                 & $64\times64\times512$                 \\
                                         & Conv                               & $3\times3/1$                 & $64\times64\times256$                 \\
                                         & Conv                               & $3\times3/1$                 & $64\times64\times128$                 \\
                                         & Conv                               & $3\times3/1$                 & $64\times64\times64$                  \\ \bottomrule
    \end{tabularx}
    \caption{\textbf{Detailed network structure.} The modules in the global branch of the mapping network are highlighted in gray.}
    \label{table:network}
\end{table}

\subsection{Defect Region Detection} \label{method:defect}
% \todo{It is weird to place detection part here. It can be briefed in the implementation.}
Since the global branch of our restoration network requires a mask $m$ as the guidance, in order to get the mask automatically, we train a scratch detection network in a supervised way by using a mixture of real scratched dataset and synthetic dataset. 
%Given synthetic image set $\cS \subset \Rbb^{H\times W \times 3}$ with the associated segmentation maps $\cY\subset [0,1]^{H\times W}$, where $H$ and $W$ denote height and width respectively. 
Specifically, let $\{s_i,y_i|s_i\in \cS,y_i\in \cY\}$ denote the whole training pairs, where $s_i$ and $y_i$ are the scratched image and the corresponding binary scratch mask respectively, %We train a network $\cF_\theta$ parameterized by $\theta$, to predict the probability of local defects at each location, thus obtaining the predicted segmentation map $\hat{y_i} = \cF_\theta(s_i)$. 
we use the cross-entropy loss to minimize the difference between the predicted mask $\hat{y_i}$ and $y_i$, 
\begin{equation}
    \label{eq:cross_entropy}
    \begin{split}
        \cL_{CE} = & \Ebb_{(s_i,y_i)\sim(\cS,\cY)}\ \Bigg\{\alpha\sum_{h=1}^{H}\sum_{w=1}^{W} - y_i^{(h,w)} \log \hat{y_i}^{(h,w)} \\
        & - (1-\alpha)\sum_{h=1}^{H}\sum_{w=1}^{W}(1-y_i^{(h,w)}) \log (1-\hat{y_i}^{(h,w)})\Bigg\}.
    \end{split}
\end{equation}
Since the scratch regions are often a small portion of the whole image, here we use a weight $\alpha_i$ to remedy the imbalance of positive and negative pixel samples. To determine the detailed value of $\alpha_i$, we compute the positive/negative proportion of $y_i$ on the fly,
\begin{equation}
    \alpha_i = \frac{[y_i=1]}{[y_i=1] + [y_i=0]]}.
\end{equation}
Besides, we also introduce the focal loss to focus on the hard samples,
\begin{equation}
    \label{eq: FL}
    \cL_{FL} = \Ebb_{(s_i,y_i)\sim(\cS,\cY)}\Bigg\{\sum_{h=1}^{H}\sum_{w=1}^{W}-(1-p_i^{(h,w)})^\gamma  \log p_{i}^{(h,w)} \Bigg\},
\end{equation}
where,
\begin{equation}
    p_i^{(h,w)}=\left\{\begin{array}{ll}{\hat{y_i}^{(h,w)}} & {\text { if } y_i^{(h,w)}=1} \\ {1-\hat{y_i}^{(h,w)}} & {\text { otherwise }}\end{array}\right.
\end{equation}
Therefore, the whole detection objective is
\begin{equation}
    \label{eq: detection_whole}
    \cL_{Seg}=\cL_{CE}+\beta\cL_{FL}.
\end{equation}
By default, we set the parameters in Equations~\eqref{eq: FL} and \eqref{eq: detection_whole} with $\gamma=0.2$ and $\beta=10$. And the detection network adopts U-Net architecture which reuses low-level features through extensive skip connection.

\subsection{Face Enhancement}
\begin{figure}[tb!]
    \begin{center}
        \includegraphics[width=1.0\linewidth]{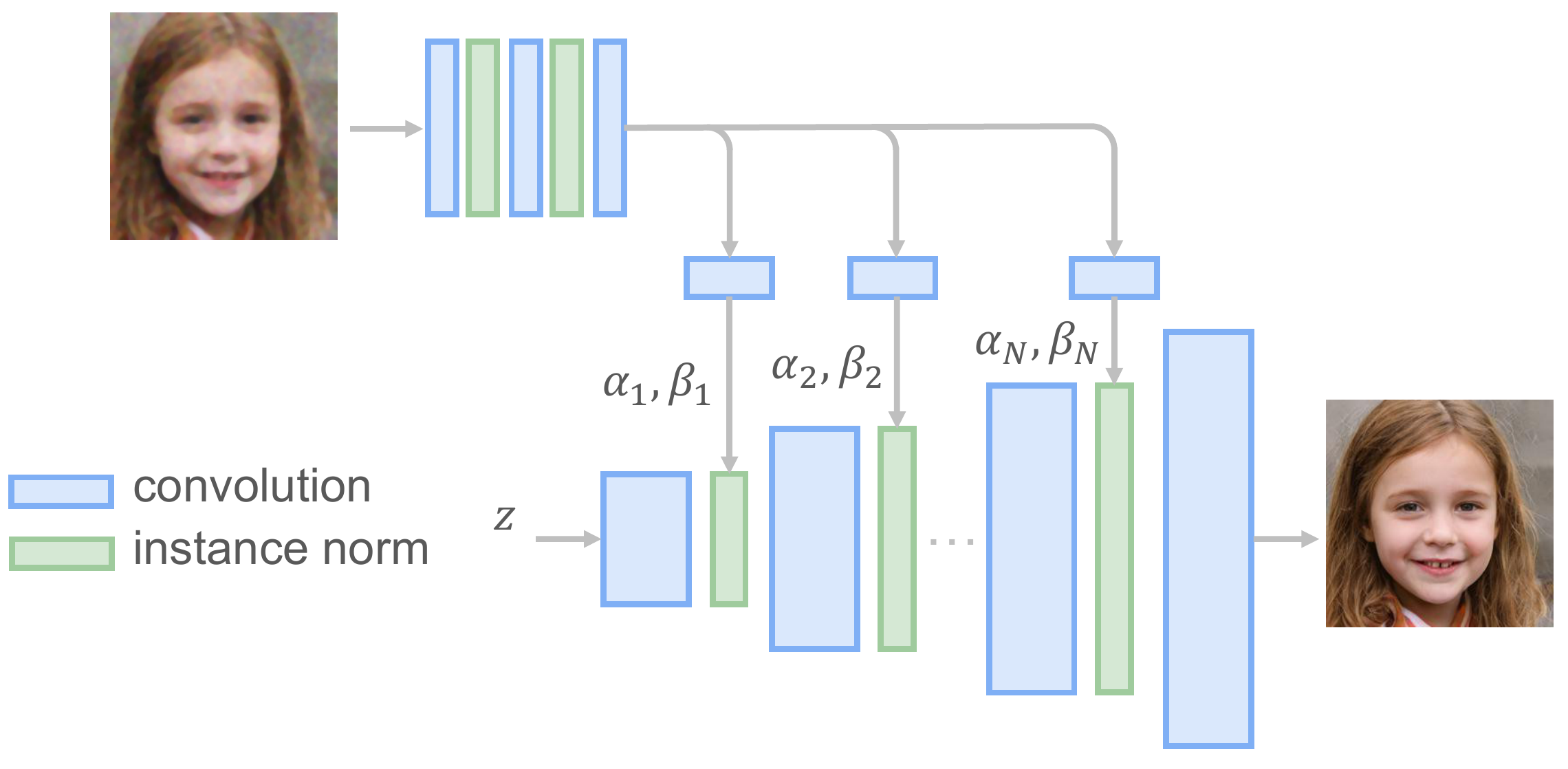}
    \end{center}
    \caption{\textbf{The progressive generator network of face enhancement.} Starting from a latent vector $z$, the network up-samples the feature map by deconvolution progressively. The degraded face will be injected into different resolutions in a spatial condition manner.}
    \label{figure:face_G}
\end{figure}
The restoration network proposed above is general to all kinds of old photos. However, considering restoration quality on faces is most sensitive to human perception, we further propose a network for face enhancement. Given one real degraded photo $r$, we hope to reconstruct degraded faces $r_f$ in $r$ into a much detailed and clean version with proposed face enhancement network $G_f$. The classical pixel-wise translation method could not solve such a blind restoration problem well because the degradation prior is totally unknown. Here, we solve this problem from the perspective of generative models.

As shown in Figure~\ref{figure:face_G}, we employ a coarse-to-fine generator to translate a low-dimensional code $z$ into corresponding high-resolution and clean faces, where $z$ is a down-sampled patch of $r_f$ ($8 \times 8$ in our implementation). At the same time of progressive generation, $r_f$ will be injected into the generator in each scale with a spatial adaptive manner~\cite{park2019semantic}, which could capture the style and structure information of degraded faces as much as possible. Specifically, let $h \in \mathbb{R}^{ H \times W \times C}$ be the activation map of previous layer and $r_f^{i}$ be the condition of current scale $i$. $h$ will be modulated as follows,
\begin{equation}
    \gamma_{x, y, c}(r_f^{i}) \frac{h_{ x, y, c}-\mu_{c}}{\sqrt{\sigma_{c}^{2}+\epsilon}}+\beta_{x, y, c}(r_f^{i}),
\end{equation}
where $h_{ x, y, c}$ denotes each element of $h$, $x\in H$ and
$y\in W$ span spatial dimensions and $c \in C$ is the feature channel. $\mu_{c}$ and $\sigma_{c}$ are the mean and standard deviation of
the activation $h$ in channel $c$. $\epsilon$ is a constant factor to avoid outlier values, $\gamma_{x, y, c}(r_f^{i})$ and $\beta_{x, y, c}(r_f^{i})$ are two learnable scalars locally controlling the influence from $r_f^{i}$. In practice, we use two convolutional layers to generate these two coefficients at each element location.

    To train the proposed face enhancement network, we penalize the perceptual distance between the generated face $G_f(z,r_f)$ and high-resolution $r_c$ as follows,
\begin{equation}
\mathcal{L}_{\text{perc}}^{\text{face}}=\Ebb\ \Big[ \sum_i \frac 1 {n_{\text{VGG}}^i} \| \phi_{\text{VGG}}^i (G_f(z,r_f)) - \phi_{\text{VGG}}^i (r_c)\|_1\Big],
\end{equation}
    where $r_f$ is the degraded face of $r_c$ and $z$ is the latent code of $r_f$. Besides, another adversarial loss is involved in the training procedure to ensure the synthesis of high-frequency details,
    
\begin{equation}
    \begin{split}
        \cL_{\text{GAN}}^{\text{face}}(z,r_f,r_c) =\ &\Ebb_{z \sim \cZ, r_f\sim \cR_f}[D_{f}(G_f(z,r_f))^2]\\ &+ \Ebb_{r_c\sim \cR_c}[(1-D_{f}(r_c))^2].
    \end{split}
\end{equation}

The face enhance network is jointly trained with previous restoration network to ensure better generalization ability, \ie, $r_f$ is the output of triplet domain translation network. We found such training scheme could effectively suppress the generated artifacts. More detailed analysis about joint training could be found in Section \ref{sec: FE_ablation}.  During inference, we firstly search the face parts of arbitrary photos, and then refine this region with proposed enhancement network. As a result of generative model, there sometimes exists color shifting between reconstructed faces and input degraded faces. We solve this issue by histogram matching. Finally the reconstructed face will be combined with original input photo using linear blending to produce the final results.

\section{Experiment}
\subsection{Implementation}
\label{sec:implementation}
\noindent\textbf{Training Dataset}~~
We synthesize old photos using images from the Pascal VOC dataset~\cite{everingham2015pascal}. In Section~\ref{sec:data_generation}, we introduce how to render realistic defects. Besides, we collect 5,718 old photos to form the images old photo dataset. To train the face enhancement network, we use
50,000 aligned high-resolution face images from FFHQ~\cite{karras2019style}.

\begin{figure}[t!]
    \begin{center}
        \includegraphics[width=0.9\linewidth]{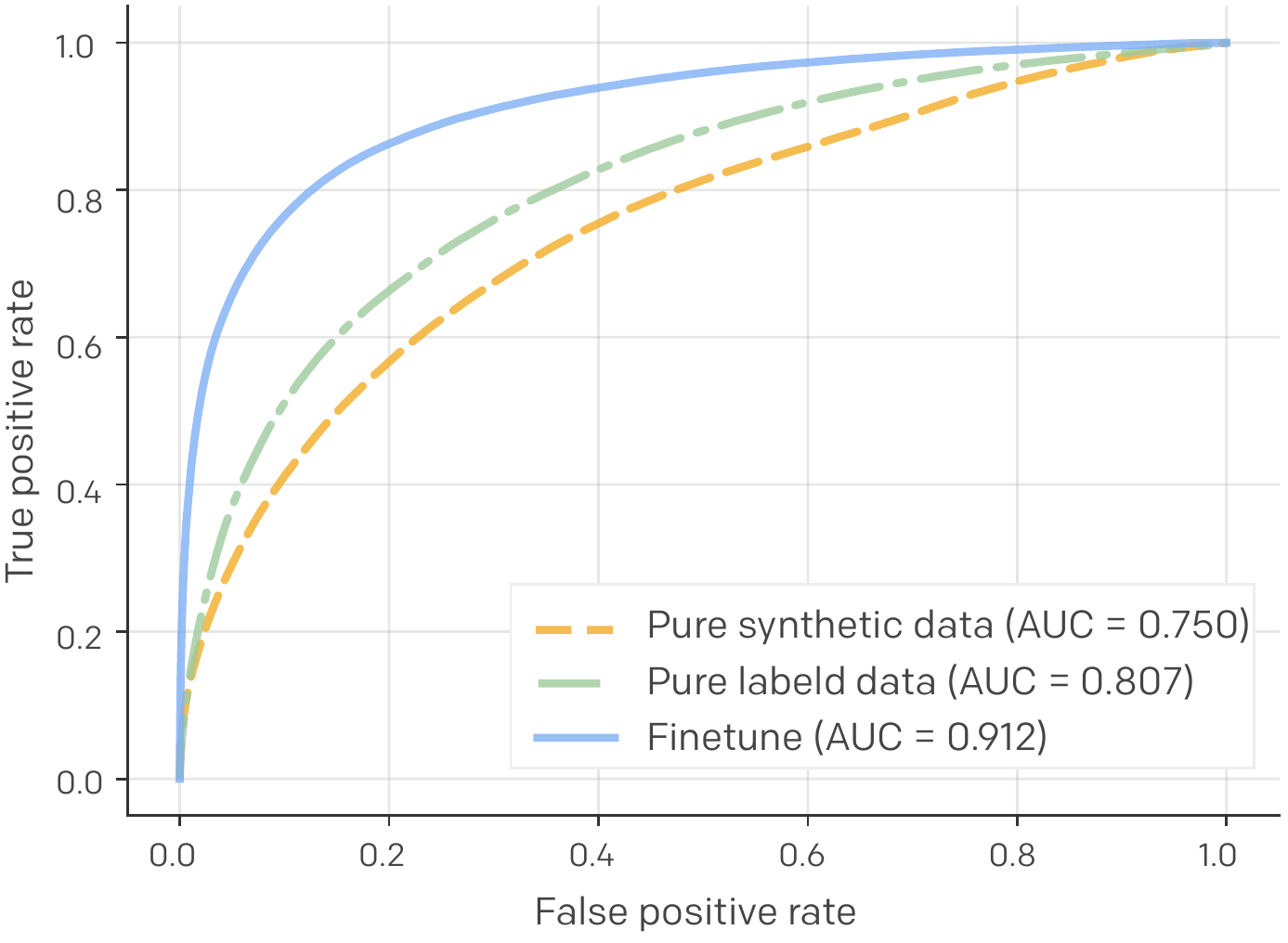}
    \end{center}
    % \vspace{-1em}
    \caption{\textbf{ROC curve for scratch detection of different data settings.} Combining both synthetic structured degradations and a small amount of labeled data, the scratch detection network could achieve great results.}
    \label{figure:roc_curve}
    % \vspace{-1.5em}
\end{figure}

% \vspace{0.3em}
% \noindent\textbf{Defect region detection}~~

% To detect structured area for the parital nonlocal block, We train another network with U-Net architecture~\cite{ronneberger2015u}. The detection network is first trained using the synthetic images only. We adopt the focal loss~\cite{lin2017focal} to remedy the imbalance of positive and negative detections. To further improve the detection performance on real old photos, we annotate 783 collected old photos with scratches, among which we use 400 images to finetune the detection network.  During inference, we apply multiscale testing. The ROC curves on the validation set in Figure~\ref{figure:roc_curve} show the effectiveness of finetuning. The area under the curve (AUC) after finetuning reaches 0.91.

% \vspace{0.3em}
\noindent\textbf{Training details}~~
We adopt Adam solver~\cite{kingma2014adam} with $\beta_1=0.5$ and $\beta_2=0.999$. The learning rate is set to 0.0002 for the first 100 epochs, with linear decay to zero thereafter.  During training, we randomly crop images to 256$\times$256. In all the experiments, we empirically set the parameters in Equations~\eqref{eq:vae1} and \eqref{eq:translation} with $\alpha=10$, $\lambda_1=60$ and $\lambda_2=10$ respectively.
% We conduct all the experiments with 4 NVIDIA Tesla P100 GPUs.

\subsection{Data Generation}\label{sec:data_generation}
\revise{Next, we brief the old photo synthesis procedure. Though we cannot fully emulate the old photo style, a careful synthesis is vital to high-quality restoration as support overlap between two domain distributions eases domain adaptation~\cite{kumar2020understanding}.} 

\noindent\textbf{Unstructured Degradation}~~
We use the following operations to simulate the unstructured degradation. Specifically, 
\begin{enumerate}
    \item Gaussian
    white noise with $\sigma \in(5,50)$. 
    \item Gaussian blur with kernel size $k\subset \{3,5,7\}$ and standard
    deviation $\sigma \in(1.0,5.0)$;
    \item JPEG compression
    whose quality level in the range of $(40,100)$;
    \item Color jitter which randomly shifts the RGB color channels by $(-20,20)$; 
    \item Box blur to mimic the lens defocus. 
\end{enumerate}
    We apply above types of augmentations with varying
    parameters in random order. To achieve more variations, we stochastically drop out each type of operation with 30\% probability. 
    Still, the synthesis cannot exactly match the appearance of real photo defects, thus requiring the proposed network to further reduce the domain gap. 

\begin{figure*}[!t]
    \begin{center}
        \small
        \begin{tabularx}{1.0\linewidth}{YYYYYYYY}
            Input & DIP & CycleGAN & Sequential & Operation-wise attention & Pix2Pix & Ours w/o face refinement & Ours w/ face refinement
        \end{tabularx}
        \includegraphics[width=1\linewidth]{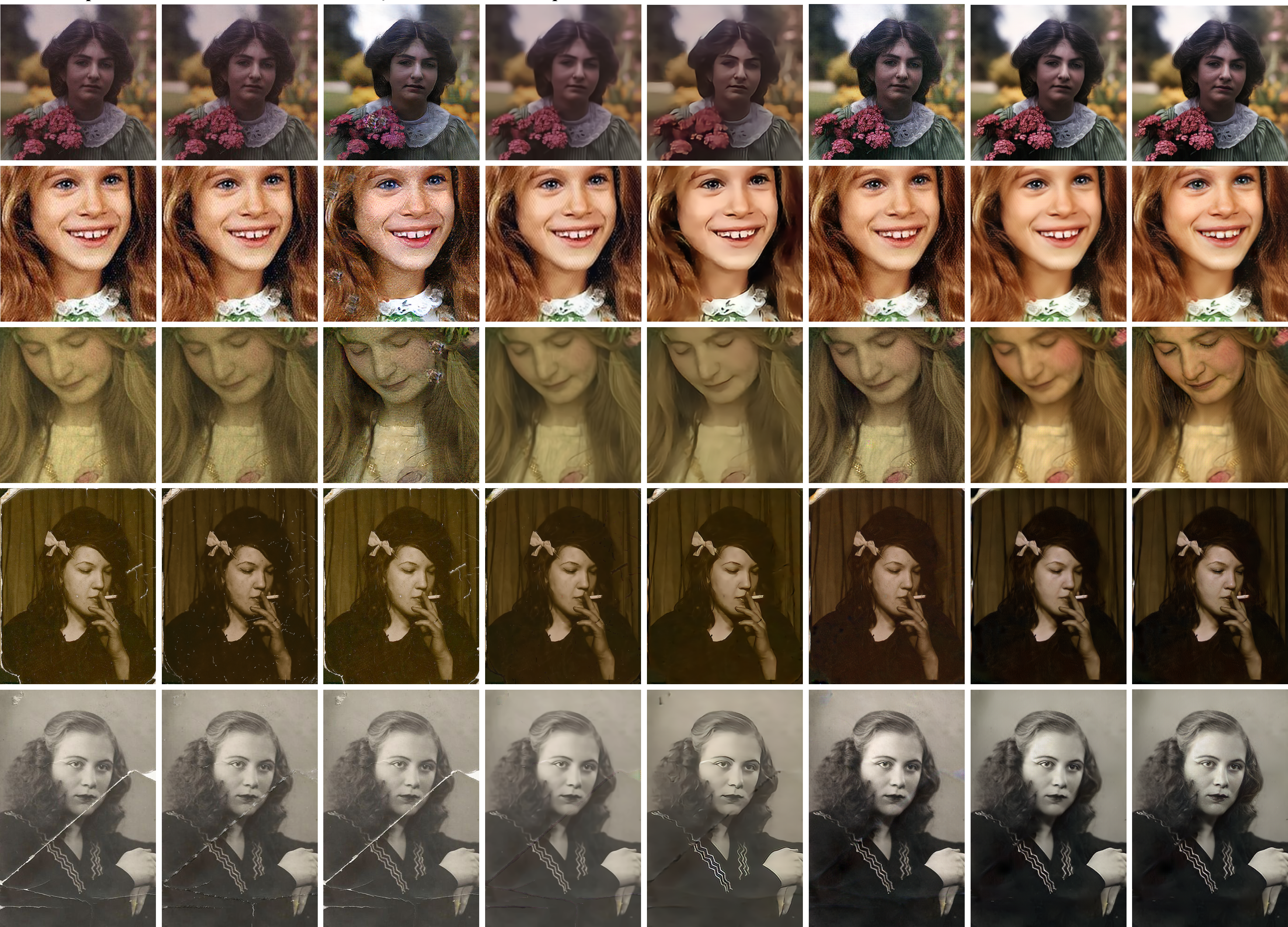}
    \end{center}
    % \vspace{-0.8em}
    \caption{\textbf{Qualitative comparison against state-of-the-art methods.} It shows that our method can restore both unstructured and structured degradation and our recovered results are significantly better than other methods.}
    \label{figure:final_comparison}
    % \vspace{-0.6em}
\end{figure*}

\begin{figure}[!tb]
    \begin{center}
        \includegraphics[width=1.0\linewidth]{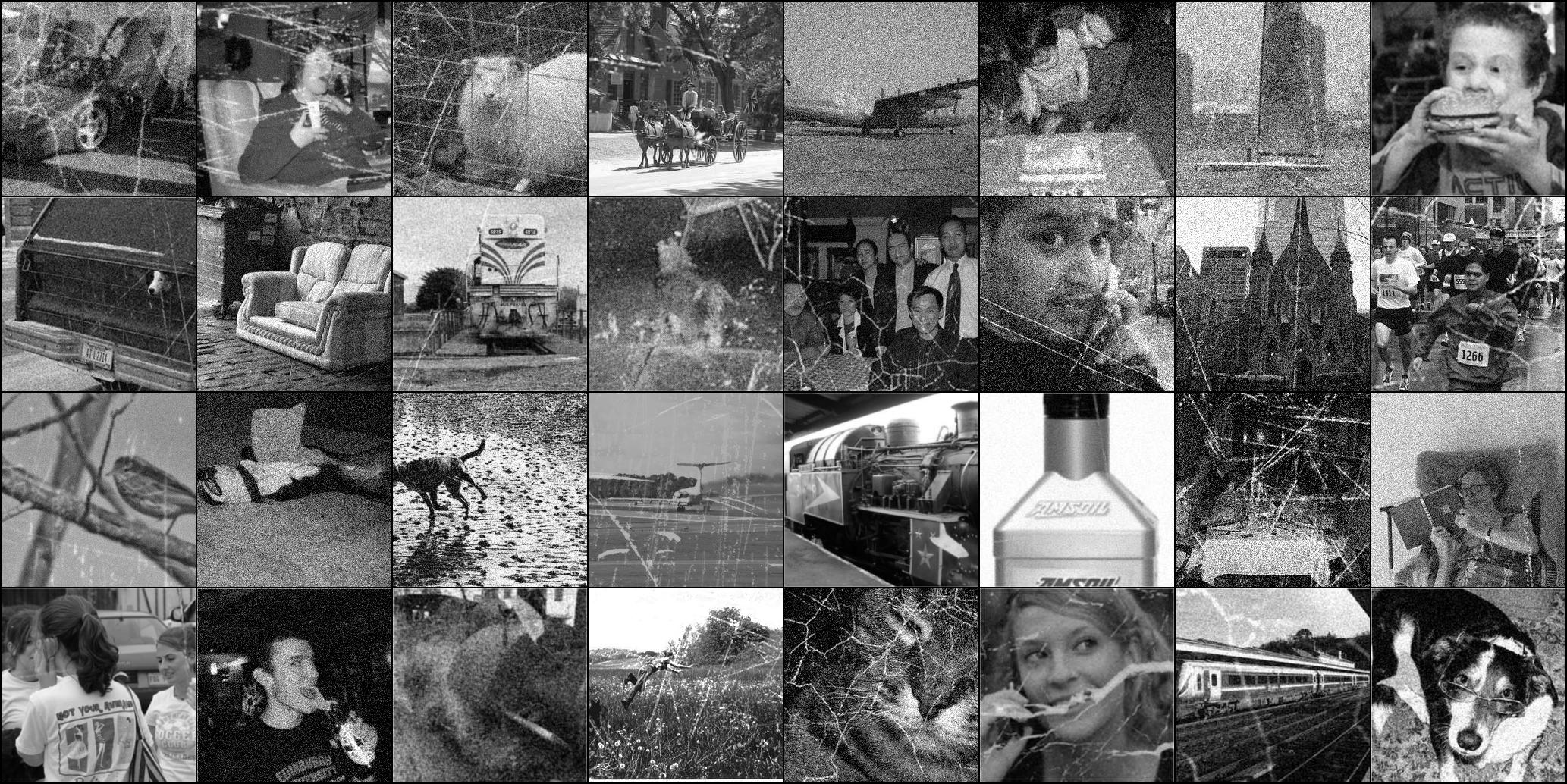}
    \end{center}
    \caption{\textbf{Some examples of synthetic photos with scratches.}}
    \label{figure:synthetic_scratch}
\end{figure}

\noindent\textbf{Structured Degradation}~~ As described in Section \ref{method:defect}, to train the defect region detection network, a mixture of synthetic and real scratch datasets are used (pretrain on synthetic and finetune on real). For the synthetic part, we collect 62 scratch texture images and 55 paper texture images, which are further augmented with elastic distortions. Then we use layer addition, lighten-only and screen modes with random level of opacity to blend the scratch textures over the natural images from the Pascal VOC dataset~\cite{everingham2015pascal}. Besides, in order to simulate large-area photo damage, we generate holes with feathering and random shape where the underneath paper texture is unveiled. Note that we also introduce film grain noise and blur with random kernel to simulate the global defects at this stage so that the synthetic data has a similar global style as the real old photos. These injected noises are beneficial in that they make the distribution of synthetic and real data become more overlapped. Examples of synthesized scratched old photos are shown in Figure~\ref{figure:synthetic_scratch}. 

\begin{figure}[!tb]
    \begin{center}
        \includegraphics[width=1.0\linewidth]{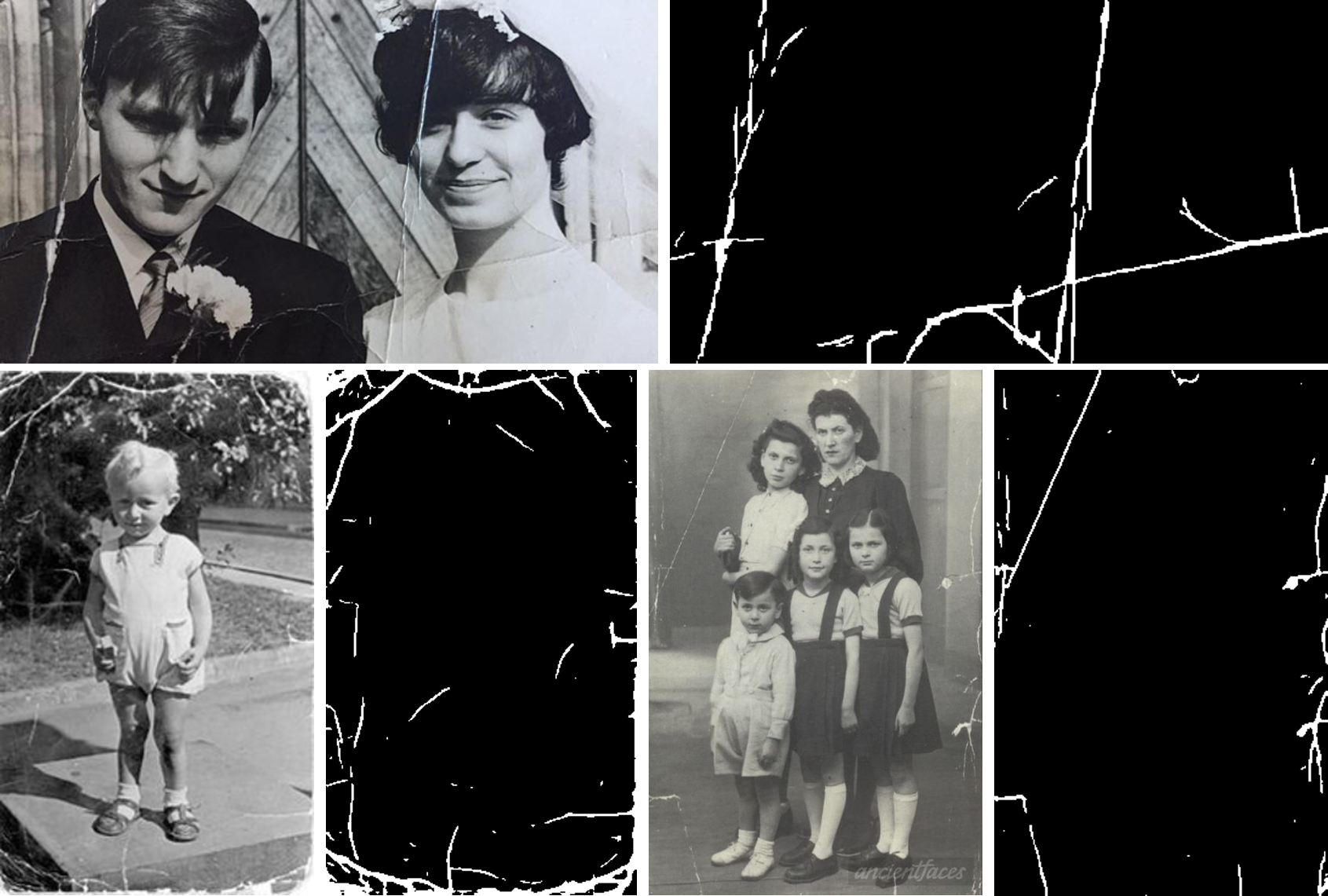}
    \end{center}
    \caption{\textbf{Some defect region detection results on real photos.}}
    \label{figure:mask}
\end{figure}

To improve the detection performance on real old photos, we collect 783 real old photos and manually annotate the local defects, among which 400 images are used for training and remaining for testing.As shown in Figure \ref{figure:roc_curve}, adding the real data into training can significantly boost the scratch detection performance on real old photos and achieve AUC as $0.912$.  Some sampled scratch detection masks and restoration results of test dataset are shown in Figure~\ref{figure:mask}.

\subsection{Comparisons}
% \vspace{0.3em}
\noindent\textbf{Baselines}~~
We compare our method against state-of-the-art approaches. For fair comparison, we train all the methods with the same training dataset (Pascal VOC) and test them on the corrupted images synthesized from DIV2K dataset~\cite{Agustsson_2017_CVPR_Workshops} and the test set of our old photo dataset. The following methods are included for comparison.

\begin{itemize}[leftmargin=*]
    \itemsep0em
    \item Operation-wise attention~\cite{suganuma2018attention} performs multiple operations in parallel and uses an attention mechanism to select the proper branch for mixed degradation restoration. It learns from synthetic image pairs with supervised learning.
    \item Deep image prior~\cite{ulyanov2018deep} learns the image restoration given a single degraded image, and has been proven powerful in denoising, super-resolution and blind inpainting.
    \item Pix2Pix~\cite{pix2pixhd} is a supervised image translation method, which leverages synthetic image pairs to learn the translation in image level.
    \item CycleGAN~\cite{CycleGAN} is a well-known unsupervised image translation method that learns the translation using unpaired images from distinct domains.
    \item The last baseline is to sequentially perform  BM3D~\cite{dabov2009bm3d}, a classical denoising method, and EdgeConnect~\cite{nazeri2019edgeconnect}, a state-of-the-art inpainting method, to restore the unstructured and structured defects respectively.
\end{itemize}

% \vspace{0.3em}
\noindent\textbf{Quantitative comparison}~~
We test different models on the synthetic images from DIV2K dataset and adopt four metrics for comparison. Table~\ref{table:quantitative_ablation_partial_nl} gives the quantitative results. The peak signal-to-noise ratio (PSNR) and the structural similarity index (SSIM) are used to compare the low-level differences between the restored output and the ground truth. The operational-wise attention method unsurprisingly achieves the best PSNR/SSIM score since this method directly optimizes the pixel-level $\ell_1$ loss. Our method ranks second-best in terms of PSNR/SSIM. However, these two metrics characterizing low-level discrepancy, usually do not correlate well with human judgment, especially for complex unknown distortions~\cite{zhang2018perceptual}. Therefore, we also adopt the recently learned perceptual image patch similarity (LPIPS)~\cite{zhang2018perceptual} metric which calculates the distance of multi-level activations of a pretrained network and is deemed to better correlate with human perception. This time, Pix2pix and our method give the best scores with a negligible difference. The operation-wise attention method, however, shows inferior performance under this metric, demonstrating it does not yield good perceptual quality. Besides, we adopt Fr\'echet Inception Distance (FID)~\cite{FID} which is a widely used metric for assessing the quality of generative models. Specifically, the FID score calculates the distance between the feature distributions of the final outputs and the real images. Still, our method and Pix2pix rank the best, while our method shows a slight quantitative advantage. In all, our method is comparable to the leading methods on synthetic data.

\begin{table}[!tb]
    \small
    \begin{center}
        \setlength{\tabcolsep}{1.3mm}{
            \begin{tabular}{@{}lcccc@{}}
                \toprule
                Method                                                & PSNR $\uparrow$ & SSIM $\uparrow$ & LPIPS $\downarrow$ & FID $\downarrow$         \\ \midrule
                Input                                                 & 12.92           & 0.49            & 0.59               & 306.80                   \\
                Attention~\cite{suganuma2018attention}                & \textbf{24.12}  & \textbf{0.70}   & 0.33               & 208.11                   \\
                DIP~\cite{ulyanov2018deep}                            & 22.59           & 0.57            & 0.54               & 194.55                   \\
                Pix2pix~\cite{pix2pixhd}                              & 22.18           & 0.62            & \textbf{0.23}      & \textbf{135.14}          \\
                % CycleGAN~\cite{CycleGAN} & --& --& --& --\\
                Sequential~\cite{dabov2009bm3d,nazeri2019edgeconnect} & 22.71           & 0.60            & 0.49               & 191.98                   \\
                \midrule
                Ours w/o partial nonlocal                                           & 23.14           & 0.68            & 0.26               & {143.62}                 \\
                Ours w/ partial nonlocal                                            & \textbf{23.33}  & \textbf{0.69}   & \textbf{0.25}      & \textbf{\textbf{134.35}} \\ \bottomrule
            \end{tabular}
            \caption{\textbf{Quantitative results on the DIV2K dataset.} Upward arrow ($\uparrow$) indicate that a higher score denotes a good image quality. We highlight the best two scores for each measure.}
            \label{table:quantitative_ablation_partial_nl}
        }
        % \vspace{-2em}
    \end{center}
\end{table}

\begin{figure*}[t!]
    \begin{center}
        \small
        \begin{tabularx}{0.9\linewidth}{YYYY}
            Input & Remini & Meitu & Ours
        \end{tabularx}
        \includegraphics[width=0.9\linewidth]{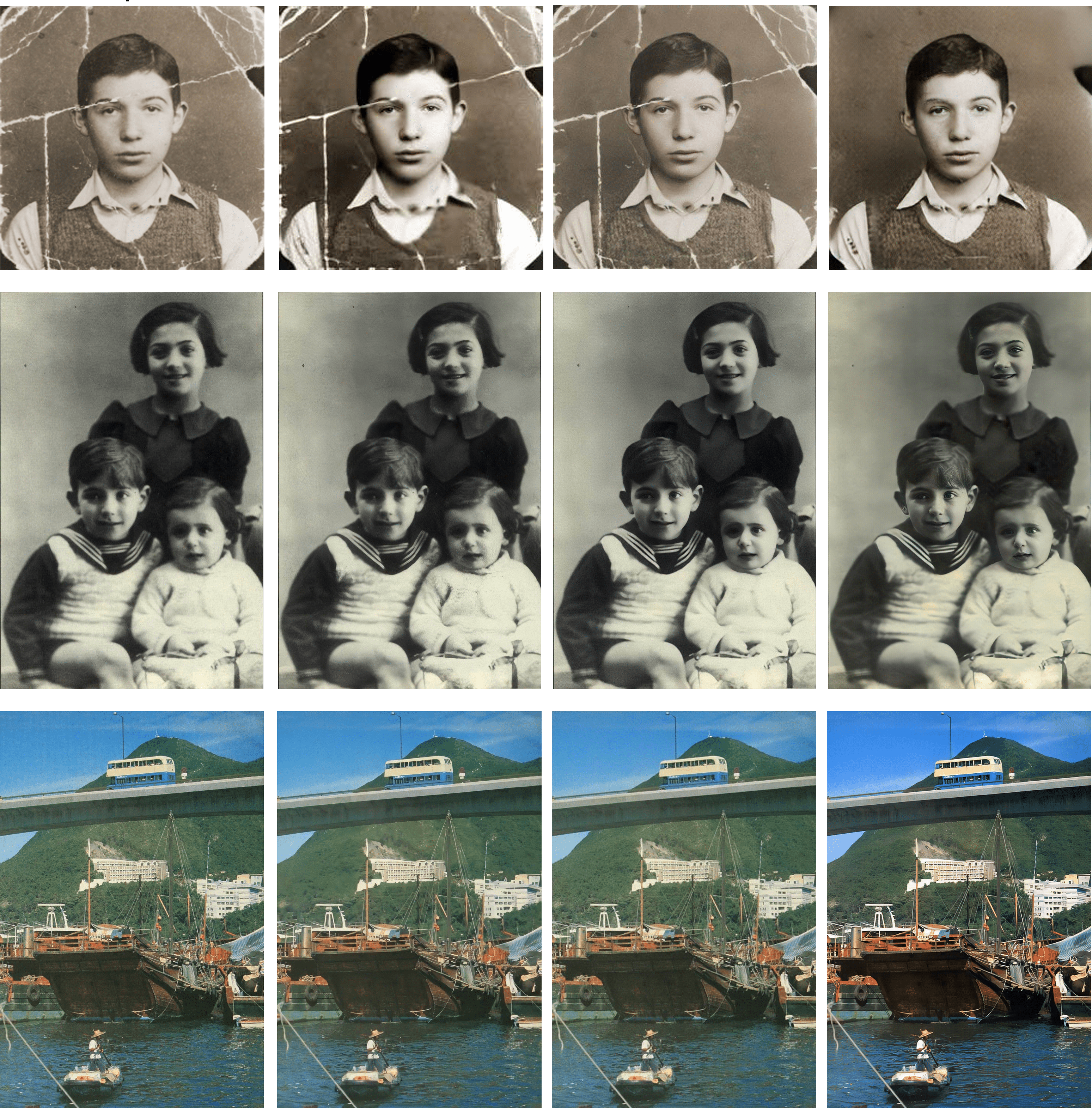}
    \end{center}
    % \vspace{-1em}
    % \caption{\textbf{Qualitative comparisons against commercial tools.} Remini Photo Enhancer~\cite{niwo}, Meitu~\cite{meitu} and our full pipeline results are included.\todo{another face result? more landscape result? more colorful result as our method can enhance color? the spacing between images is too large.}}
    \caption{\textbf{Qualitative comparisons against commercial tools.} Remini Photo Enhancer~\cite{niwo}, Meitu~\cite{meitu} and our full pipeline results are included.}
    \label{figure:qualitative_full_comparison}
    %\vspace{-0.6em}
\end{figure*}

\begin{figure*}[t!]
    \begin{center}
        \small
        \begin{tabularx}{1.0\linewidth}{YYYYY}
            Input & Pix2Pix & VAE & VAE-TS & Full model
        \end{tabularx}
        \includegraphics[width=1.0\linewidth]{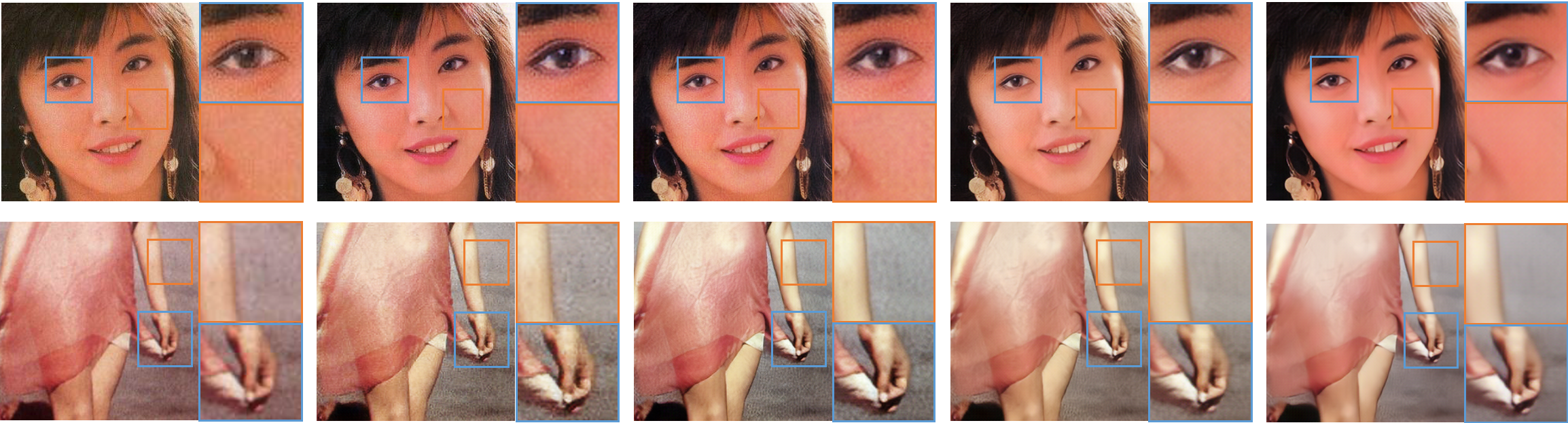}
    \end{center}
    \caption{\textbf{Ablation study for latent translation with VAEs.} By involving feature translation and feature-level adversarial loss, the domain gap between synthetic degradations and real-world defects could be narrowed gradually, leading to better restoration results step by step.}
    \label{figure:qualitative_ablation_feature_translation}
\end{figure*}

\begin{figure*}[t!]
    \begin{center}
        \small
        \begin{tabularx}{1.0\linewidth}{YYYYY}
            Input & Pix2Pix & Operation-wise attention & Ours w/o partial nonlocal & Ours w/ partial nonlocal
        \end{tabularx}
        \includegraphics[width=1.0\linewidth]{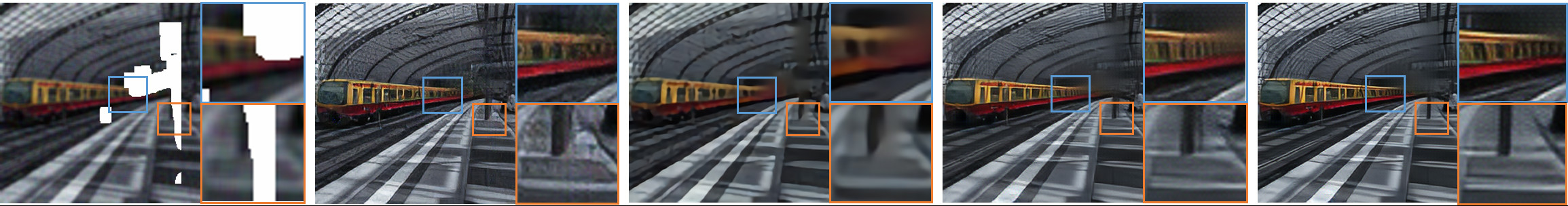}
    \end{center}
    %\vspace{-1.4em}
    \caption{\textbf{Ablation study of partial nonlocal block.} Partial nonlocal better inpaints the structured defects.}
    \label{figure:qualitative_ablation_PN}
    %\vspace{-0.8em}
\end{figure*}

% \vspace{0.3em}
\noindent\textbf{Qualitative comparison}~~
To prove the generalization to real old photos, we conduct experiments on the real photo dataset. For a fair comparison, we retrain the CycleGAN to translate real photos to clean images. Since we lack the restoration ground truth for real photos, we cannot apply reference-based metrics for evaluation. Therefore, we qualitatively compare the results, which are shown in Figure~\ref{figure:final_comparison}. The DIP method can restore mixed degradations to some extent. However, there is a trade off between the defect restoration and the structural preservation: more defects reveal after a long training time while fewer iterations induce the loss of fine structures. CycleGAN, learned from unpaired images, tends to focus on restoring unstructured defects and neglect to restore all the scratch regions. Both the operation-wise attention method and the sequential operations give comparable visual quality. However, they cannot amend the defects that are not covered in the synthetic data, such as sepia issue and color fading. Besides, the structured defects still remain problematic, possibly because they cannot handle the old photo textures that are subtly different from the synthetic dataset. Pix2pix, which is comparable to our approach on synthetic images, however, is visually inferior to our method. Some film noises and structured defects still remain in the final output. This is due to the domain gap between synthetic images and real photos, which makes the method fail to generalize. In comparison, our method gives clean, sharp images with the scratches plausibly filled with unnoticeable artifacts. Besides successfully addressing the artifacts considered in data synthesis, our method can also enhance the photo color appropriately. In general, our method gives the most visually pleasant results and the photos after restoration appear like modern photographic images.

\noindent\textbf{User study}~~
To better illustrate the subjective quality, we conduct a user study to compare with other methods. We randomly select 21 old photos from the test set and let users sort the results according to the restoration quality. We collect subjective opinions from 24 users and count the percentage of each method being selected as the top $K$ ($K=1-5$). The results are shown in Table~\ref{table:user_study}, which clearly demonstrates the  advantage of our approach, with 65.43\% chances to be selected as the top 1.

\begin{table}[!tb]
    \small
    \begin{center}
        \setlength{\tabcolsep}{0.5mm}{
            \begin{tabularx}{\columnwidth}{@{}lYYYYY@{}}
                \toprule
                Method                                                & Top 1          & Top 2          & Top 3          & Top 4          & Top 5          \\ \midrule
                DIP~\cite{ulyanov2018deep}                            & 2.54           & 8.49           & 19.26          & 39.09          & 74.22          \\
                CycleGAN~\cite{CycleGAN}                              & 4.24           & 8.21           & 19.54          & 28.32          & 50.42          \\
                Sequential~\cite{dabov2009bm3d,nazeri2019edgeconnect} & 4.81           & 18.13          & 47.87          & 79.60          & 94.61          \\
                Attention~\cite{suganuma2018attention}                & 6.79           & 21.24          & 49.85          & 73.08          & 88.38          \\
                Pix2Pix~\cite{pix2pixhd}                              & 16.14          & 60.90          & 73.65          & 86.68          & 94.90          \\
                \textbf{Ours}                                         & \textbf{65.43} & \textbf{83.00} & \textbf{89.80} & \textbf{93.20} & \textbf{97.45} \\ \bottomrule
            \end{tabularx}
            \caption{\textbf{User study results.} The percentage (\%) of each method being selected as the top $K$ ($K=1-5$) by users.}
            \label{table:user_study}
        }
        %\vspace{-2em}
    \end{center}
\end{table}

\noindent\textbf{Comparison with Commercial Software}~~
\revise{
    Some commercial software and applications like Meitu~\cite{meitu} and Remini Photo Enhancer~\cite{niwo} start to provide the service of automatic old photos restoration. To demonstrate the effectiveness of our pipeline, we also compare the restoration performance with them in Figure~\ref{figure:qualitative_full_comparison}. Based on the observation of their outputs, it could be found that their methods ignore the existing structured degradations and color fading. By contrast, our method alleviates these problems and generates more visual-pleasant results like the first row and third row of Figure~\ref{figure:qualitative_full_comparison}. Meanwhile, the proposed latent domain translation network better deals with real-world local defects such as noise because of a smaller domain gap in the second row of Figure~\ref{figure:qualitative_full_comparison}. Finally, with a dedicated face enhancement network, the refined human face also contains more details. Overall, our method could achieve more clean, sharp and vibrant results compared with commercial counterparts. }

\subsection{Analysis}
\revise{
    In order to prove the effectiveness of individual technical contributions, we perform the following ablation study.
}
\subsubsection{Latent translation with VAEs}
\revise{
    Let us consider the following variants, with proposed components added one-by-one: }
\begin{itemize}[leftmargin=*]
    \itemsep0em

    \item Pix2Pix which learns the translation in image-level. The model is trained using only synthetic pairs.
    \item Two VAEs with an additional KL loss to penalize the latent space. The VAEs and latent mapping are all trained simultaneously.
    \item VAEs with two-stage training (VAEs-TS): the two VAEs are first trained separately and the latent mapping is learned thereafter with the two fixed VAEs, which ensure the translation is performed in two fix domains.
    \item Full model, which also adopts latent adversarial loss.

\end{itemize}
\revise{
    We first calculate the Wasserstein distance~\cite{arjovsky2017wasserstein} between the latent space of old photos and synthetic images. Table~\ref{table:quantitative_ablation_feature_translation} shows that distribution distance gradually reduces after adding each component. The main reason is that the KL loss of VAEs could lead to a more compact latent space. Training with the two-stage manner isolates the two VAEs, and the latent adversarial loss further closes the domain gap. A smaller domain gap will improve the model generalization to real photo restoration. To demonstrate this point, several visual comparison results (without any face post-processing) are provided in Figure~\ref{figure:qualitative_ablation_feature_translation}. We observe that Pix2Pix could not handle these blind distortions well. The restoration is gradually improved with a more compact latent space. Besides, we also adopt a blind image quality assessment metric, BRISQUE~\cite{mittal2012no}, to measure photo quality after restoration. The BRISQUE score in Table~\ref{table:quantitative_ablation_feature_translation} progressively improves by applying these mentioned techniques, which is consistent with corresponding visual results.}

\begin{table}[t!]
    \small
    \begin{center}

        \setlength{\tabcolsep}{0.5mm}{
            \begin{tabularx}{\columnwidth}{@{}lYYYY@{}}
                \toprule
                Method                   & Pix2Pix & VAEs   & VAEs-TS & full model      \\ \midrule
                Wasserstein $\downarrow$ & 1.837   & 1.048  & 0.765   & \textbf{0.581}  \\
                BRISQUE $\downarrow$     & 25.549  & 23.949 & 23.396  & \textbf{23.016}
                \\ \bottomrule
            \end{tabularx}
            % %\vspace{0.5em}
            \caption{\textbf{Ablation study of latent translation with VAEs.} We provide some quantitative comparisons here to demonstrate the superior performance of the full model. Our method achieve best results on both distribution distance and BRISQUE metric.}
            \label{table:quantitative_ablation_feature_translation}
        }
    \end{center}
    %\vspace{-1em}
\end{table}

\subsubsection{Partial nonlocal block}
\begin{figure}[!tb]
    \begin{center}
        \small
        \begin{tabularx}{1.0\linewidth}{YYYY}
            Input & Mask & w/o partial nonlocal & w/ partial nonlocal
        \end{tabularx}
        \includegraphics[width=1.0\linewidth]{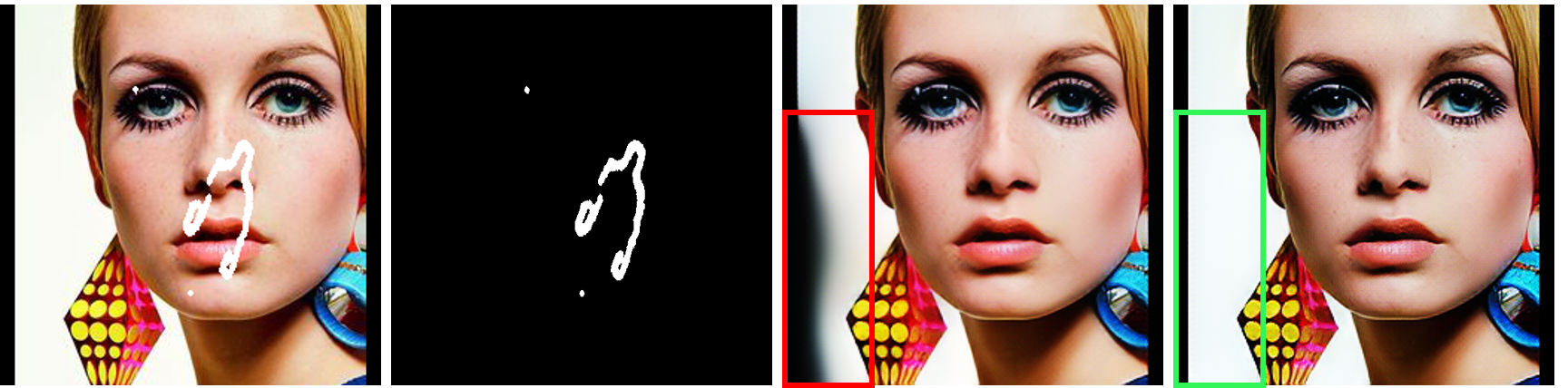}
    \end{center}
    %\vspace{-1.4em}
    \caption{\textbf{Ablation study of partial nonlocal block.} Partial nonlocal does not touch the non-hole regions as this operation is aware of the corruption area.}
    \label{figure:qualitative_ablation_PN_2}
    %\vspace{-0.8em}
\end{figure}

\revise{
  We propose the partial nonlocal block to make the triplet domain translation network support the restoration of structured degradations. As shown in Figure~\ref{figure:qualitative_ablation_PN}, both Pix2Pix~\cite{isola2017image} and mixed-distortion restoration method~\cite{suganuma2018attention} could not simultaneously handle local and global defects well. Because of the utilization of large image context (partial nonlocal), the scratches can be inpainted with fewer visual artifacts and better globally consistent restoration can be achieved in our method. In addition, we find that the partial nonlocal block could also ensure that the inpainting is only applied in the localized defect areas. In Figure~\ref{figure:qualitative_ablation_PN_2}, the background of origin photos will be modified if we remove this block. Besides, the quantitative result in Table~\ref{table:quantitative_ablation_partial_nl} also shows that the partial nonlocal block consistently improves the restoration performance on all the metrics.}

\begin{figure*}[tb!]
    \begin{center}
        \small
        \begin{tabularx}{1.0\linewidth}{YYYYY}
            Original face & Output of stage 1 & Pix2Pix & Stage 2 w/o joint training & Stage 2 w/ joint training
        \end{tabularx}
        \includegraphics[width=1.0\linewidth]{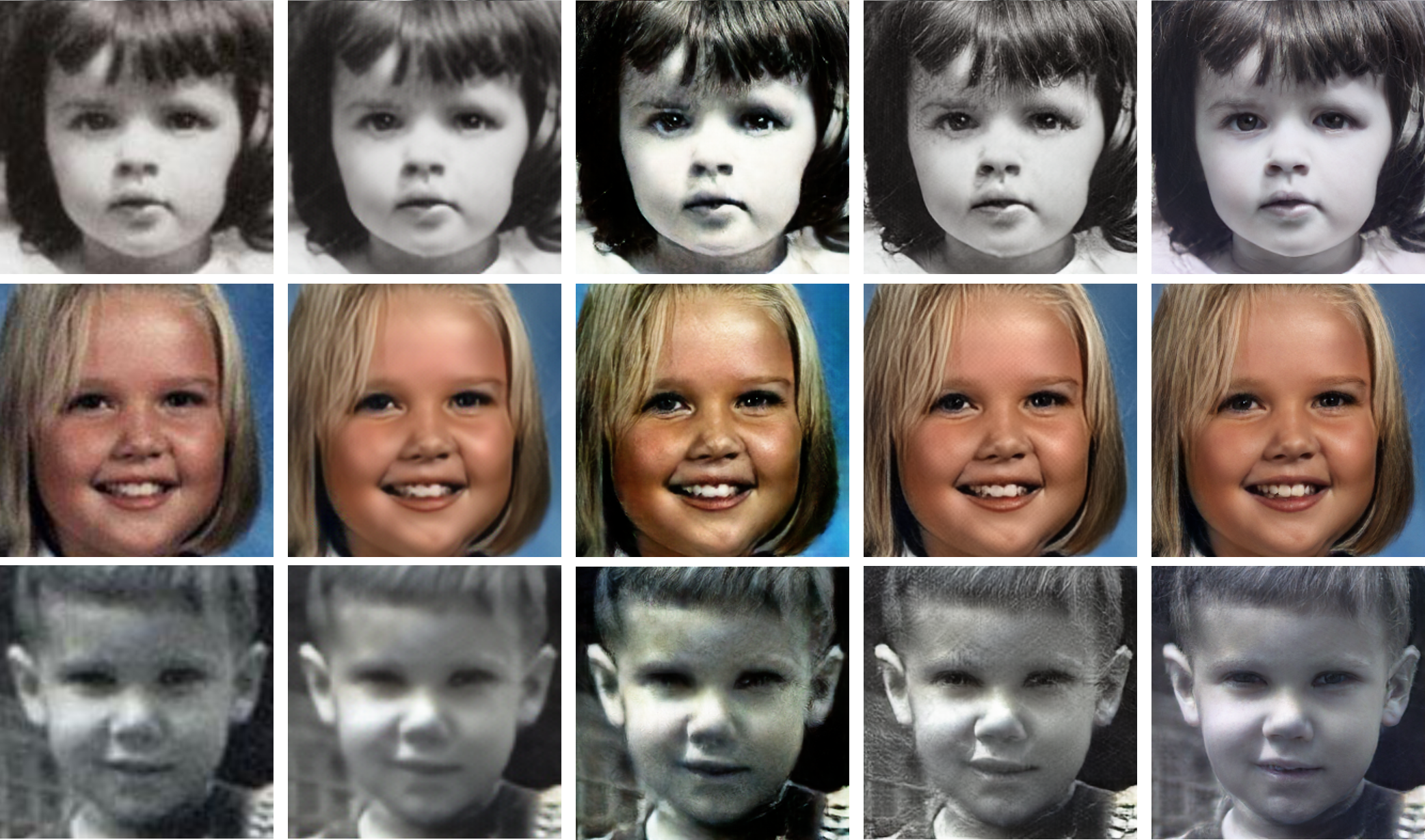}
    \end{center}
    % \caption{\textbf{Comparisons with Pix2Pix~\cite{isola2017image} and w/o joint training.}\todo{swap the pix2pix and stage 1 columns; try to find better examples; 1st and 3rd rows are not good}}
    \caption{\textbf{Comparisons with Pix2Pix~\cite{isola2017image} and w/o joint training.} To ensure fair comparison, the Pix2Pix model is also trained jointly with the domain translation network. Stage 1: Triplet domain translation. Stage 2: Face enhancement. }
    \label{figure:qualitative_ablation_joint_training}
\end{figure*}

\begin{figure*}[tb!]
    \begin{center}
        \small
        \begin{tabularx}{1.0\linewidth}{YYYYYYY}
            Original face & 16$\times$16 & 32$\times$32 & 64$\times$64 & 128$\times$128 & 256$\times$256 & Hierachical
        \end{tabularx}
        \includegraphics[width=1.0\linewidth]{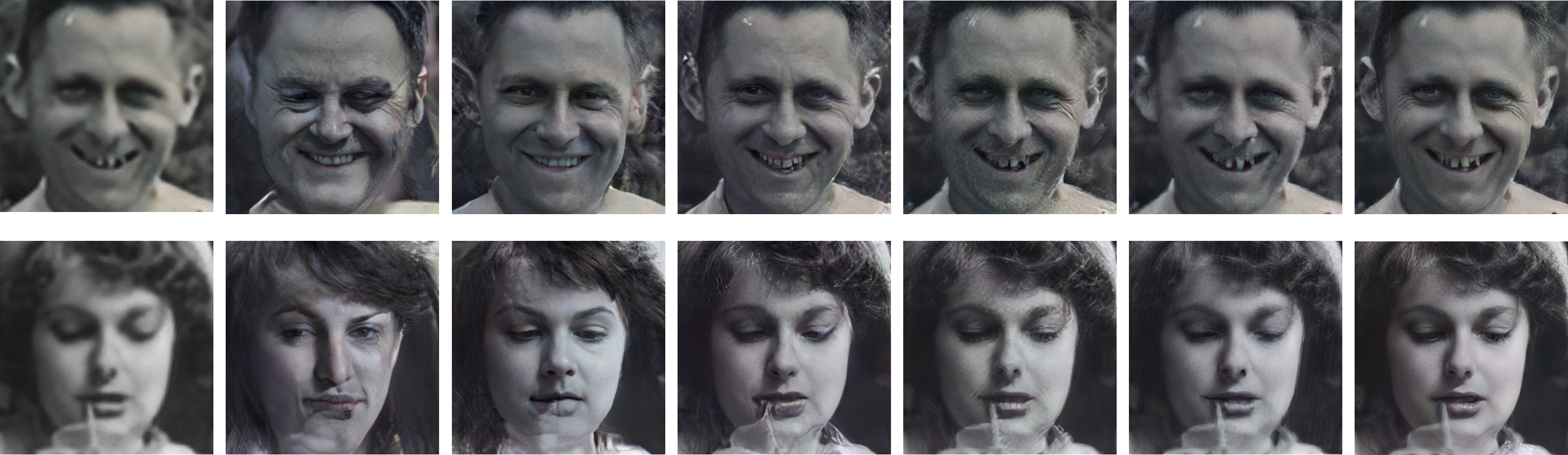}
    \end{center}
    \caption{\textbf{Face reconstruction results with different injection methods.} The hierarchical manner leads to the best results.}
    \label{figure:qualitative_ablation_hier}
\end{figure*}

\begin{table*}[t!]
    \small
    \begin{center}
        \setlength{\tabcolsep}{2.0mm}{
            \begin{tabular}{@{}lccccccc@{}}
                \toprule             & {Input} & {16 x 16} & {32 x 32} & {64 x 64} & {128 x 128} & {256 x 256} & {Hierarchical}   \\
                \midrule
                {PSNR} $\uparrow$    & 22.918  & 17.677    & 20.931    & 23.088    & 24.622      & 24.938      & \textbf{25.282} \\
                {SSIM} $\uparrow$    & 0.655   & 0.545     & 0.618     & 0.677     & 0.724       & 0.740       & \textbf{0.743}  \\
                {FID} $\downarrow$   & 42.421  & 24.177    & 17.993    & 15.768    & 14.236      & 15.653      & \textbf{13.175} \\
                {LPIPS} $\downarrow$ & 0.376   & 0.271     & 0.193     & 0.150     & 0.129       & 0.133       & \textbf{0.120}  \\
                \bottomrule
            \end{tabular}
            \caption{\textbf{Quantitative comparisons for different injection positions.} We test the results on synthetic degraded face images.}
            \label{table: quantitative_ablation_hier}
        }
        % \vspace{-2em}
    \end{center}
\end{table*}

\subsubsection{Ablation Study of Face Enhancement Network}\label{sec: FE_ablation}

\noindent\textbf{Effectiveness of Joint Training}~~
\revise{
    The face enhancement network is jointly trained with the triplet domain translation network, i.e., input corrupted faces will first pass through this translation network, and then be reconstructed into a high-resolution version with the proposed enhancement network. To demonstrate the effectiveness of this training scheme, we provide some qualitative results of real old photos in Figure~\ref{figure:qualitative_ablation_joint_training}. We could observe that without joint training, unnatural redundant textures and artifacts are visible in the generated faces. One reason may be there exist some distribution bias between generated faces of the first stage and real degraded faces. By introducing the joint training, this kind of gap could be alleviated, leading more pleasant and stable results.}

\noindent\textbf{Comparison with Other Generative Model}~~
\revise{
      Generally, a straightforward question could arise here: if a simple pixel-wise translation model like~\cite{chan2019everybody} could obtain desired results? To verify this point, we train another Pix2Pix~\cite{isola2017image} model using the same loss function and discriminator. There are mainly two differences between Pix2Pix~\cite{isola2017image} and our enhancement network: 1)  We adopt the progressive generator with the spatial condition rather than the image-level concatenation of Pix2Pix~\cite{isola2017image}. 2) The model parameter amount of Pix2Pix~\cite{isola2017image} is about 188.9M, which is almost twice as much as ours (92.1M). The visual comparisons could be found in Figure~\ref{figure:qualitative_ablation_joint_training}. Obviously, the results of~\cite{isola2017image} are far below our expectations, which contains many artifacts and color degradations. By contrast, the reconstructed faces of our method are more vivid. The potential reason for this phenomenon is that the face post-processing of~\cite{chan2019everybody} is to replenish slight details of faces, but our target here is to reconstruct the face based on the corrupted observation, which is a more challenging task. By the feature-level spatial modulation, the generator learns to reconstruct a clean face while capturing the original structure and style information.}

\noindent\textbf{Effectiveness of Hierarchical Spatial Injection}~~
\revise{
    To reconstruct a high-resolution face from real photos meanwhile maintaining underlying structure and style information, we propose to modulate the features of the coarse-to-fine generator in a hierarchical spatial condition manner. To demonstrate the importance of this point, we compare this method with the single spatial injection of different layers, i.e., from the lowest scale ($16\times16$) to the highest ($256\times256$) one. Qualitatively, as shown in Figure~\ref{figure:qualitative_ablation_hier}, although we could generate a more vivid face at the lowest scale, the identity is not preserved since a low-dimensional condition could not constrain the generator well. With the increase of injection resolution, the reconstructed face becomes more accurate gradually. However, we find that the generated faces contain lots of noise and artifacts when the injection is performed at the highest scale only. The reason may be that the position of highest scale injection is too close to the generator output and less relevant with the semantic feature in previous layers, thus resulting in the incomplete modulation. By contrast, our hierarchical spatial injection achieves natural restoration results with the right structures and styles, as shown in the last column of Figure~\ref{figure:qualitative_ablation_hier}. To further prove this point, we also calculate the quantitative statistics of each scale on a synthetic dataset. We randomly select 2,000 test images and add varying degradations to construct paired data. As shown in Table~\ref{table: quantitative_ablation_hier}, although scale $16\times16$ and $32\times32$ achieve better performance on FID and LPIPS compared with input which demonstrates the distribution of generated face become close to real HR faces, the PSNR and SSIM are even lower than the input because of the loss on original information. By introducing the method of hierarchical injection, our enhancement network obtains the best scores on all four metrics.}

\setlength{\tabcolsep}{1.0pt}
\begin{figure}[t!]
    \begin{center}
        \small
        \begin{tabularx}{1.0\columnwidth}{YYYY}
            Input & Ours & Input & Ours
        \end{tabularx}
        \includegraphics[width=1.0\linewidth]{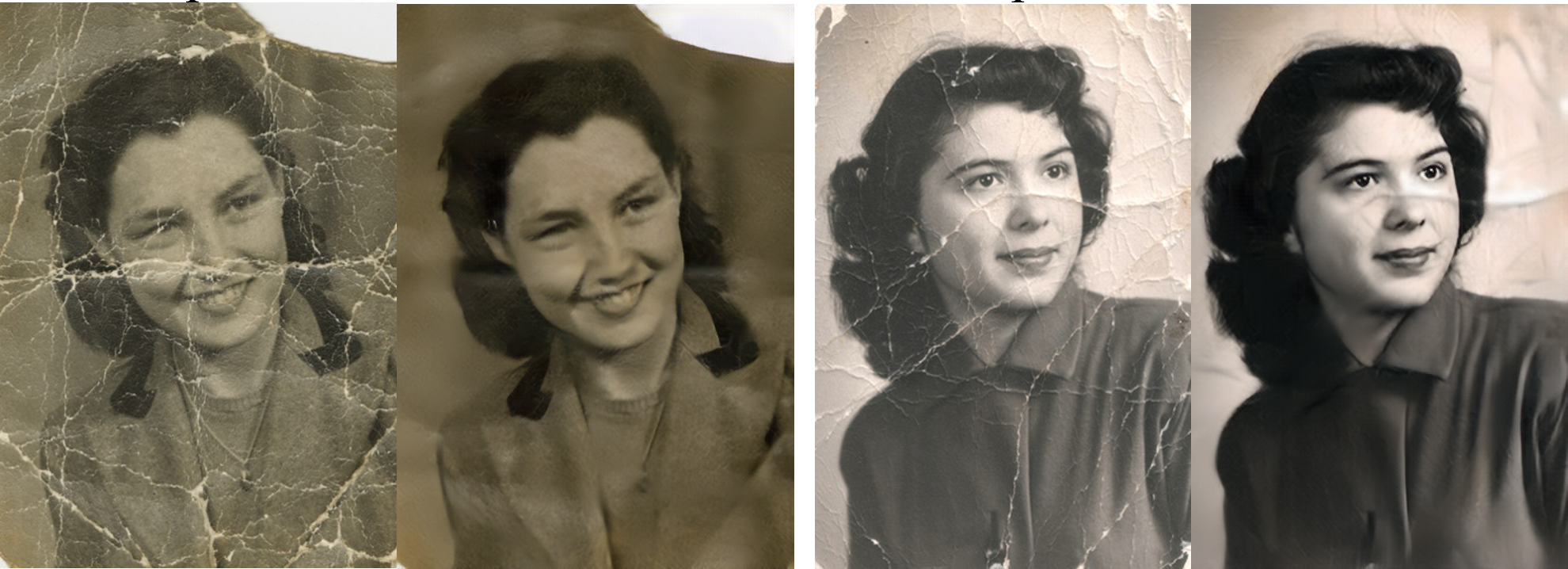}
    \end{center}
    %\vspace{-1.4em}
    \caption{\textbf{Limitation.} Our method cannot handle complex shading artifacts due to the uneven lighting.}
    \label{figure:limitation}
    %\vspace{-1em}
\end{figure}

\section{Discussion and Conclusion}
We propose a novel triplet domain translation network that opens new avenue to restore the mixed degradation for in-the-wild old photos. The domain gap is reduced between old photos and synthetic images, and the translation to clean images is learned in latent space. Our method suffers less from generalization issue compared with prior methods. Besides, we propose a partial nonlocal block which restores the latent features by leveraging the global context, so the scratches can be inpainted with better structural consistency. Furthermore, we propose a coarse-to-fine generator with spatial adaptive condition to reconstruct the face regions of old photos. Our method demonstrates good performance in restoring severely degraded old photos. However, our method cannot handle complex shading as shown in Figure~\ref{figure:limitation}. This is because our dataset contains few old photos with such defects. One could possibly address this limitation using our framework by explicitly considering the shading effects during synthesis or adding more such photos as training data.

\ifCLASSOPTIONcaptionsoff
    \newpage
\fi

% trigger a \newpage just before the given reference
% number - used to balance the columns on the last page
% adjust value as needed - may need to be readjusted if
% the document is modified later
%\IEEEtriggeratref{8}
% The "triggered" command can be changed if desired:
%\IEEEtriggercmd{\enlargethispage{-5in}}

% references section

% can use a bibliography generated by BibTeX as a .bbl file
% BibTeX documentation can be easily obtained at:
% http://mirror.ctan.org/biblio/bibtex/contrib/doc/
% The IEEEtran BibTeX style support page is at:
% http://www.michaelshell.org/tex/ieeetran/bibtex/
\bibliographystyle{IEEEtran}
% argument is your BibTeX string definitions and bibliography database(s)
\bibliography{Restoration}
\end{document}